\newcommand{\indep}{\perp \!\!\! \perp}

\newcommand{\vZ}{\bm{Z}}
\newcommand{\vz}{\bm{z}}
\newcommand{\vr}{\bm{r}}
\newcommand{\vR}{\bm{R}}

\newcommand{\Ni}[1]{\mathcal{N}_i^{#1}}
\newcommand{\szo}{\{0,1\}}
\newcommand{\Gr}{\mathcal{G}}
\newcommand{\Gri}{\Gr_i^n}

\newcommand{\Ed}{\mathcal{E}}
\newcommand{\Exp}{\mathbbm{E}}
\newcommand{\Pro}{\mathbbm{P}}

\documentclass[mnsc,nonblindrev]{informs3} 
\usepackage{bm}

\usepackage{amsmath}
\usepackage{algpseudocode}
\usepackage{algorithm}
\usepackage{CJKutf8}
\usepackage{changes}
\usepackage{amssymb}
\usepackage{bbm}
\usepackage{amsmath}
\usepackage{graphicx}
\usepackage{enumitem}
\usepackage{setspace}
\usepackage{threeparttable}
\usepackage{multirow}
\usepackage{pdflscape}
\usepackage{accents}
\newcommand{\ubar}[1]{\underaccent{\bar}{#1}}
\usepackage{mathtools}
\usepackage{booktabs}
\usepackage{array}

\newcommand{\prf}[1]{\textit{Proof. } #1 \hfill}

\usepackage{setspace}

\OneAndAHalfSpacedXI

\usepackage{natbib}
 \bibpunct[, ]{(}{)}{,}{a}{}{,}%
 \def\bibfont{\small}%
 \def\bibsep{\smallskipamount}%

\TheoremsNumberedThrough
 \EquationsNumberedThrough  
 \MANUSCRIPTNO{MS-XXXX}

\begin{document}

\RUNTITLE{Network Interference Heterogeneity}

\TITLE{A Two-Part Machine Learning Approach to Characterizing Network Interference in A/B Testing}
 
 \ARTICLEAUTHORS{%
\AUTHOR{Yuan Yuan}
\AFF{Graduate School of Management, University of California, Davis, CA, 95616}
\AUTHOR{Kristen M. Altenburger}
\AFF{Central Applied Science, Meta Inc., Menlo Park, CA, 94025}
}

\ABSTRACT{The reliability of controlled experiments, commonly referred to as ``A/B tests," is often compromised by network interference, where the outcomes of individual units are influenced by interactions with others. Significant challenges in this domain include the lack of accounting for complex social network structures and the difficulty in suitably characterizing network interference. To address these challenges, we propose a machine learning-based method. We introduce ``causal network motifs'' and utilize transparent machine learning models to characterize network interference patterns underlying an A/B test on networks. Our method's performance has been demonstrated through simulations on both a synthetic experiment and a large-scale test on Instagram. Our experiments show that our approach outperforms conventional methods such as design-based cluster randomization and conventional analysis-based neighborhood exposure mapping. Our approach provides a comprehensive and automated solution to address network interference for A/B testing practitioners. This aids in informing strategic business decisions in areas such as marketing effectiveness and product customization.
}


\maketitle

\section{Introduction}
Controlled experiments, also known as ``A/B testing,'' continue to serve as the cornerstone for making strategic decisions in business, including new product launches, marketing campaigns, and algorithm updates~\citep{bakshy2014designing,kohavi2020trustworthy,bojinov2022online,koning2022experimentation}. 
Through the random assignment of treatment or control groups, A/B testing facilitates the evaluation of  causal,  rather than merely correlational, impacts of a product intervention on business outcomes. 
Businesses have increasingly recognized the value of A/B testing and are investing in the development of in-house experimentation platforms~\citep{kohavi2013online,bakshy2014designing,xu2015infrastructure}.
Furthermore, numerous companies have adopted A/B testing software like Optimizely and Split to efficiently perform and analyze their A/B tests. Companies adopting A/B testing have seen performance improvements of 30\% to 100\% within a year~\citep{koning2022experimentation}.

However, a significant obstacle to the validity of A/B testing is network interference. Conventional causal inference rests on an essential assumption known as the ``Stable Unit Treatment Value Assumption" (SUTVA)~\citep{rubin2005causal}, 
which implies a unit's outcome only depends on their treatment assignment. 
Network interference occurs when a unit's (e.g., a person's) outcome is influenced by the treatment assignments of other units, especially those within their network neighborhood, if their connections are modeled as a network~\citep{hudgens2008toward,toulis2013estimation,basse2018model}.
Network interference is prevalent in numerous contemporary A/B testing environments, including social media, online marketplaces, and location-based platforms~\citep{hagiu2015marketplace,yan2018strategic,holtz2020reducing,li2022interference}.  Failing to properly account for network interference is problematic. For instance, \cite{holtz2020reducing} found that network interference could skew the estimation of the treatment effect by more than 30\%. Overall, network interference presents a significant challenge to industrial A/B tests, as it may mislead business decisions on product updates if the test results are unreliable.

Efforts to improve estimation in the presence of network interference have predominantly followed two research paths: pre-experimental design and post-experiment analysis~\citep{eckles2016design}. The focus of pre-experimental design is to devise improved randomization treatment assignments. A notable instance is graph cluster randomization~\citep{ugander2013graph}, a variant of cluster randomization~\citep{bland2004cluster}, which establishes graph clusters (communities) based on the graph structure and assigns treatments at the level of these graph clusters.
Works such as those by \cite{saveski2017detecting} and \cite{pouget2019testing} have introduced strategies for verifying the existence of network interference and demonstrated the superiority of graph cluster randomization over conventional Bernoulli randomization in the presence of network interference. \cite{ugander2020randomized} further enhanced the cluster randomization design by suggesting a randomized graph clustering approach. More recently, \cite{viviano2020experimental} 
and \cite{candogan2023correlated} 
proposed approaches to minimize the variance of the estimators related to graph cluster randomization.
Concurrent research has also explored specific applications where network interference occurs. {A sizable portion of contemporary research, for instance, concentrates on experimental designs for bipartite networks such as two-sided marketplaces~\citep{bajari2021multiple,zigler2021bipartite,brennan2022cluster,johari2022experimental,harshaw2023design}. Moreover, when geographic units are conceptualized as nodes and physical proximity as network links, it is referred to as ``spatial interference"~\citep{pollmann2020causal,wang2020design}.
Recently, the research spotlight has increasingly centered on switchback experiments, which randomize across time rather than at the individual level, representing another type of design-based approach~\citep{bojinov2022design,cortez2022graph,boyarsky2023modeling}.

As for post-experiment analysis methods, many existing methods can be understood through the \textit{exposure mapping} framework~\citep{aronow2017estimating} (or equivalently, effective treatments by \cite{manski2013identification}). Exposure mapping describes the dependency structure between an individual unit's outcome and the treatment assignments of other units. Within this framework, interference is modeled by establishing exposure conditions, each of which depends not only on the treatment assignment of the individual unit but also on the treatment assignments or attributes of other units.
A simple example of exposure mapping is the \textit{fractional $q$ neighborhood exposure}, which categorizes units based on whether the proportion of treated neighbors exceeds a specific threshold~\citep{ugander2013graph,eckles2016design}. There is an expanding body of literature that extends the conventional causal inference framework to accommodate network interference~\citep{bowers2013reasoning,eckles2016design,athey2018exact,basse2019randomization,forastiere2021identification,leung2022causal,hu2022average,yu2022estimating}.  
Moreover, recent developments in post-experiment analysis have also been tailored to specific applications, such as marketplaces~\citep{munro2021treatment} and spatial experiments~\citep{wang2021causal}.

Our study sets out to enhance post-experiment analysis by introducing a two-part framework designed to specify a class of exposure mappings for network experiments. We aim to address two key limitations in the current exposure mapping literature. First, the existing exposure mapping framework does not consider the local network structure of interference, such as the connectedness among a user's friends or tie strength. Previous studies in network science have underscored the crucial role of local network structure and tie strength for explaining user behavior~\citep{aral2014tie,kim2017strength,lyu2022investigating}. 
For instance, the structural diversity hypothesis suggests that the likelihood of product adoption heavily depends on the extent of disconnectedness among a unit's neighbors who have adopted the product~\citep{aral2011diversity,ugander2012structural}.
However, the current interference literature has rarely accounted for the local network structure of interference.

The second limitation is that specifying the exposure mapping function currently relies too heavily on human experts manually modeling the network interference structure~\citep{aronow2017estimating}. Different experimenters may propose various exposure mappings, which might not align with the true interference patterns of a specific experiment. Even with a fixed population across multiple experiments, interference patterns from past experiments may not be applicable to the current one. This makes it challenging for experimenters to define an appropriate exposure mapping for a particular experiment. Moreover, even with a predefined exposure mapping function, experimenters must rely on their judgment to set the parameters within this function, which may not yield ideal results.

We propose a two-step approach to addressing the above limitations. 
 In the first step, we propose the concept of ``causal network motifs.'' This combines (1) network motifs~\citep{milo2002network,shen2002network} which encapsulate interaction patterns among a node and its ego network, and (2) the treatment assignment of each user as a label-dependent feature~\citep{gallagher2008leveraging}. 
 Causal network motifs are presented as a vector representation, each dimension of which mirrors the treatment conditions of a specific type of network motif.\footnote{While our focus is on undirected and unweighted networks in this paper, our approach can easily be extended to directed, weighted, or multilayer networks by constructing various types of vector representations.}
In the second step, we introduce an unsupervised learning (clustering) approach to map causal network motifs to specific exposure conditions. While most clustering algorithms can be adapted for this task, including the tree-based algorithm introduced in the prior conference version~\citep{yuan2021causal}, we focus on presenting a nearest neighbors method designed to estimate the global average treatment effect, which compares potential outcomes under conditions of universal treatment versus no treatment. Our approach offers an interpretable framework to understand patterns of interference and improve estimation. Additionally, we provide theoretical guidelines to balance bias and variance when designing exposure mapping specification algorithms.
Empirically, we tested our methods in both synthetic and real-world settings, demonstrating that our approach can reduce bias more effectively than standalone cluster randomization or fractional $q$ neighborhood exposure mapping.

\noindent 
\textbf{Running Example.} 
Consider a new feature on a social media platform that introduces a novel messaging capability, such as sending a new emoji. Practitioners aim to assess the effectiveness of this feature through an A/B test. Users randomly selected into the treatment group receive access to the new messaging feature. The primary outcomes measured include the increase in time spent on the platform and the volume of messages sent. However, there is potential for network interference, as users in the control group, who do not have access to the new feature, may also alter their behavior. This change occurs because they interact with friends who have access to the new feature, which can influence them to spend more time on the platform and engage more frequently. Our study aims to address the bias in commonly used estimators that arises from  network interference.

\section{Exposure Mapping Framework And Overview of Our Approach}
\subsection{Preliminaries}
\label{ref:preliminaries}

Consider a finite population $\mathcal{N}$ and a network denoted by $\Gr = \left(\mathcal{N}, \mathcal{E}\right)$, where $\mathcal{N}$ and $\mathcal{E}$ are node and edge sets, respectively. 
Let $|\mathcal{N}|=N$. 
Nodes or edges may have covariates.
Let $i$ (or $j$) index units in a network for $i \in \{1,...,N\}$. $\vZ=(Z_1, Z_2, \dots, Z_N)$ is a random vector in $\szo^N$ and under a specific experiment design denoted by $\Pro_{\vZ}$. The realization of this random vector is denoted by $\vz=(z_1, z_2, \dots, z_N)$.\footnote{For example, Bernoulli randomization assumes that $Z_i \overset{\mathrm{iid}}{\sim} \mathrm{Bern}(p)$ where $p \in (0, 1)$; by contrast, graph cluster randomization \citep{ugander2013graph,eckles2016design} is reflected by a different probability measure $\Pro_{\vZ}$. 
Although our study only considers binary treatments ($Z_i=1$ or $0$), this can easily extend beyond binary treatment conditions. } Note that here the treatment assignment is the sole source of randomness. 

In network settings, the unit's outcome can be dependent on the treatment assignment conditions of neighbors or even other units\footnote{Examples include social contagion or displacement \citep{aral2012identifying,yuan2019gift,weisburd2014hot}.}, which violates  the Stable Unit Treatment Value Assumption (SUTVA)~\citep{imbens2010rubin}.\footnote{This means the potential outcome $y_i(\cdot)$ does not depend on the treatment assignment of any other users (i.e., $\vZ_{-i}$).}
In the presence of interference, $y_i$ would be a mapping from $y_i: \{0,1\}^N \rightarrow \mathbb{R}$, where $\mathbb{R}$ is the set of all real numbers and 
\begin{equation*}
Y_i = \sum_{\vz \in \{0,1\}^N } \mathbbm{1}[\vZ=\vz] y_i(\vz).    
\end{equation*}
That is, each realization of the random assignment vector $\vZ$ may lead to a  completely different observed outcome $Y_i$. Therefore, there could be $2^N$ rather than $2$ potential outcomes for each unit. 

We consider each node (unit)'s $n$-hop ego networks:
\begin{definition}[$n$-Hop Ego Network]
The $n$-hop neighbor set of unit $i$ is defined by recursion: (1) $\Ni{0} = \{i\}$; (2) $\Ni{n+1} = \Ni{n} \cup \{ j  | (j,k) \in \Ed \text{ and } k \in \Ni{n} \}$. The $n$-hop ego network of $i$ is denoted by $\Gri$. $\Gri$ is the vertex-induced subgraph of $\Gr$ by the vertex set $\Ni{n}$, i.e., $\Gri = (\Ni{n}, \Ed_i^n)$ where $\Ed_i^n = \{ (j,k) | j \in \Gri,  k \in \Gri, (j,k) \in \Ed \}$.
\end{definition}
\noindent 

We adopt the neighborhood interference assumption:  interference is restricted to the $n$-hop ego network neighborhood but not beyond:

\begin{assumption}[$n$-hop Neighborhood Interference]
$y_i(\vz) = y_i(\vz')$ if for all $\vz, \vz' \in \{0,1\}^N$ such that $z_j = z'_j$ when $j \in \Ni{n}$.
\end{assumption}
\noindent  
Statistical tests can be performed to verify this assumption. \cite{athey2018exact} provide a method to compute the exact $p$-value for the hypothesis of no network interference beyond $n$ hops.
\footnote{Previous studies such as~\cite{eckles2016design,chin2019regression,cortez2022exploiting} also adopt this assumptions. \cite{leung2022causal,belloni2022neighborhood} have explored the choice of optimal $n$, which is beyond the scope of our study.} In the running example, the $1$-hop neighborhood interference implies that only the random assignments received by a user's immediate friends on social media may influence the user's outcome.

 Our characterization of interference extends the \textit{exposure mapping} framework initially proposed by \cite{aronow2017estimating}, a classical framework for addressing interference:

\begin{definition}[Exposure Mapping]
Let $f$ be an exposure mapping function $f: \{0, 1\}^N \times \Theta \rightarrow \Delta$.
Here $\theta_i \in \Theta$ describes the attribute (including ego network structure) of unit $i$ and $\Delta$ is the set of exposure conditions.
\end{definition}
The exposure mapping framework identifies multiple conditions beyond simple treatment or control by considering not only the treatment received by unit $i$  but also the assignments of their network neighbors. 
One example is the \textit{fractional $q$ neighborhood exposure mapping} \citep{ugander2013graph} (see Figure~\ref{fig:q}), which defines four exposure conditions ($|\Delta|=4$) by combining whether the ego node is treated with whether more than a fraction $q$ of its neighbors are assigned to the same treatment condition. Under this specification, an exposure condition could be, e.g., ``the ego user is treated and more than 60\% of its social media friends are also treated'' in our running example.

Under exposure mapping \(f\) and its corresponding exposure condition \(\delta\), we can define the \textit{expected potential outcome} for unit \(i\) as follows:
\begin{equation}
\mu_i^f(\delta) = \Exp_{\vZ} [ y_i(\vZ) | f(\vZ, \theta_i) = \delta].
\label{eq:exp}
\end{equation}
Intuitively, this means that under the probability distribution \(\mathbb{P}_{\mathbf{Z}}\), we are evaluating the expectation of \(Y_i\) conditional on the fact that \(i\) has been randomly assigned to an exposure condition \(\delta\).

\cite{aronow2017estimating} assume that practitioners can correctly specify an exposure mapping $f$:
\begin{definition}[Correctly Specified Exposure Mapping]
Exposure mapping $f$ is correctly specified if $\mu_i^f(\delta) = y_i(\vz)$ for all $\vz \in \{0,1\}^N$ such that $f(\vz, \theta_i) = \delta$.
\end{definition}

\cite{aronow2017estimating} left an open question on how to find an exposure mapping suitable for a given experiment and data. Our approach extends the exposure mapping framework by incorporating both network motifs and machine learning algorithms to specify an exposure mapping $f$.\footnote{Note that a subtle distinction from our framework is that we do not necessarily assume that $f$ has to be correctly specified. Instead, we adopt the perspective that while identifying a correctly specified $f$ is challenging, even a misspecified $f$ can be valuable under certain assumptions. Therefore, the estimators presented in Section~\ref{sec:estimation} remain useful under misspecification, provided that we adjust our causal estimands to be based on Eq.~\eqref{eq:exp}.}

\subsection{Estimation Tasks}
\label{sec:estimation}
Within the exposure mapping framework, there are two critical tasks: estimating the average potential outcomes and the global average treatment effects.

\noindent\textbf{Average Potential Outcomes.}
Under the exposure mapping framework, each exposure condition is associated with an average potential outcome, which represents what would occur if every individual were assigned to an exposure condition (e.g., the ego unit is treated and more than 60\% of their neighbors are also treated). Formally, for each $\delta \in \Delta$, the average potential outcome is defined as:
\[    \mu^f(\delta) = \frac{1}{N} \sum_{i\in\mathcal{N}} \mu^f_i(\delta) = \frac{1}{N} \sum_{i\in\mathcal{N}} \Exp_{\vZ} [ y_i(\vZ) | f(\vZ, \theta_i) = \delta].\]

\noindent\textbf{Global Average Treatment Effects.}
The second task is to estimate the global average treatment effect~\citep{ugander2013graph}, which compares the fully treated versus the fully non-treated counterfactual scenarios. Let us define
\begin{equation*}
\mu(\bm{1})  = \frac{1}{N} \sum_{i\in\mathcal{N}}  y_i(\bm{1}); \quad \mu(\bm{0}) = \frac{1}{N} \sum_{i\in\mathcal{N}}  y_i(\bm{0}); \quad \tau = \mu(\bm{1}) - \mu(\bm{0}),
\end{equation*}
\noindent where $\mu(\bm{1})$ and $\mu(\bm{0})$ represent the average outcomes when every unit is treated or not treated, respectively, and $\tau$ denotes the global average treatment effect.
In our running example, $\mu(\bm{1})$ indicates the average time spent on the platform (or volume of messages sent) when all users are granted access to the new messaging feature. Conversely, $\mu(\bm{0})$ measures the average time spent on the platform when no users have access.\footnote{Note that $\mu(\bm{1})$ and $\mu(\bm{0})$ do not depend on any exposure mapping and are simply the averages over all potential outcomes, whereas $\mu^f(\cdot)$ depends on an exposure mapping $f$ and is an expectation of a function of $\bm{Z}$.} The global average treatment effect, $\tau$, quantifies the overall impact of the new messaging feature on time spent on the platform by comparing the two counterfactuals—one where all users are granted access and one where none are granted access.

\noindent\textbf{General Probability of Exposure and Positivity Requirement.}
Before introducing our estimators, we first define the positivity requirement:
\begin{definition}[Positivity Requirement]
For an exposure condition $\delta \in \Delta$, the positivity requirement is satisfied if $\Pro[f(\vZ,\theta_i)=\delta]>0$ for all $i \in \mathcal{N}$.
\end{definition}
\noindent The probability $\Pro[f(\vZ,\theta_i)=\delta]$ is referred to as the \textit{general probability of exposure}. The rationale of ensuring positivity is that it allows our estimators to operate effectively with non-zero denominators in their calculations.

A key computational challenge is the estimation of this probability. We utilize Monte Carlo simulations to approximate this probability (see \textit{Appendix~\ref{sec:appendix:general}} for details). Note that Monte Carlo simulations are readily parallelizable, thus improving computational efficiency.

\noindent\textbf{Estimators.}
Here we discuss the common estimators for the two tasks. 
To estimate the average potential outcome under an exposure condition $\delta \in \Delta$,
we adopt the following estimators
\begin{equation}
\hat{\mu}^f_{\text{HT}}(\delta) =  \frac{\sum_i Y_i w^f_i(\delta)}{N} \quad\text{ and } \quad
\hat{\mu}^f_{\text{H\'ajek}}(\delta) =  \frac{\sum_i Y_i w^f_i(\delta)}{\sum_i w^f_i(\delta)}  \quad \text{ where } w^f_i(\delta) =  \frac{\mathbbm{1}[ f(\vZ,\theta_i) = \delta ]}{\mathbbm{P}[ f(\vZ,\theta_i) = \delta ]}.\label{eq:Hajek_u} 
\end{equation}

Horvitz-Thompson (HT) estimator ($\hat{\mu}^f_{\text{HT}}(\delta)$) is an unbiased estimator to its estimand (${\mu}^f(\delta)$):
\begin{lemma}
Given an exposure condition $\delta$ that satisfies the positivity requirement,
\begin{equation}
 \Exp_{\vZ}\left[\hat{\mu}^f_{\text{HT}}(\delta)\right] = {\mu}^f(\delta).
\end{equation} \label{eq:lemma:est}
\end{lemma} 
\vspace{-1cm}\prf{See \textit{Appendix~\ref{sec:lemma:est}}.}

The H\'ajek estimator, while exhibiting a slightly smaller bias, empirically demonstrates much lower variance compared to the HT estimator. Many related studies thus recommend using the H\'ajek estimator to trade a small increase in bias for substantial variance reduction~\citep{eckles2016design, aronow2017estimating, khan2021adaptive}.

Under the exposure mapping framework, two specific exposure conditions must be defined to estimate the global average treatment effects: $\delta^{(1)}$ and $\delta^{(0)}$. These conditions approximate the scenarios where all units are fully treated ($\vz = \mathbf{1}$) and fully controlled ($\vz = \mathbf{0}$), respectively. The treatment effects are then estimated as follows:
\[
\hat{\tau}_{\text{HT}} = \hat{\mu}_{\text{HT}}(\delta^{(1)}) - \hat{\mu}_{\text{HT}}(\delta^{(0)}) \quad \text{and} \quad \hat{\tau}_{\text{Hajek}} = \hat{\mu}_{\text{Hajek}}(\delta^{(1)}) - \hat{\mu}_{\text{Hajek}}(\delta^{(0)})
\]
where $\hat{\tau}_{\text{HT}}$ and $\hat{\tau}_{\text{H\'ajek}}$ are the HT and Hájek estimators of the treatment effects, respectively.

Given the two exposure conditions $\delta^{(1)}$ and $\delta^{(0)}$, we can define the estimand\footnote{Note that in later sections we discuss $\tau^f$ does not necessarily equal $\tau$ unless $f$ is correctly specified.} $$\tau^f = \mu^f(\delta^{(1)}) - \mu^f(\delta^{(0)}).$$
According to our discussion on estimators for average potential outcomes, we immediately derive $\mathbb{E}_{\mathbf{Z}}[\hat{\tau}_{\text{HT}}^{f}] = \tau^f$ and that $\mathbb{E}_{\mathbf{Z}}[\hat{\tau}_{\text{H\'ajek}}^{f}]$ has a slightly small bias when being used to estimate $\tau^{f}$.

\subsection{Overview of Our Approach}\label{ref:pipeline}

Here we provide the overview of our framework: the two-part decomposition of the exposure mapping \(f\), defined as \(f = h \cdot g\). This decomposition involves two mappings:
\begin{itemize}
    \item \(g: \{0, 1\}^N \times \Theta \rightarrow [0,1]^M\), which constructs the causal network motif representation in \([0,1]^M\);
    \item \(h: [0, 1]^M \rightarrow \Delta\), which automatically specifies the exposure condition using an unsupervised learning (clustering) algorithm.
\end{itemize}

This decomposition is detailed in Section~\ref{ref:motifs} for \(g\) and Section~\ref{ref:algo} for \(h\). The construction of \(g\) involves establishing the ``causal network motif representation." The dimensionality \(M\) is determined through a procedure of causal network motif selection.\footnote{Similar to the feature engineering process in machine learning, practitioners initially identify a wide range of features (dimensions) to characterize a unit’s network neighborhood, including treatment assignments and network structure. Subsequently, dimensionality may be reduced using methods such as Lasso regression.}
The function of \(h\) involves employing a clustering algorithm to convert the causal network motif representation into a specific exposure condition.

Our major extension of the exposure mapping framework is that we let each exposure condition $\delta \in \Delta$  correspond to a subspace $\mathcal{R} \in [0,1]^M$.\footnote{To connect $\delta$ and $\mathcal{R}$, we can introduce a mapping $\psi: \Delta \rightarrow \mathcal{P}([0,1]^M)$, where $\mathcal{P}$ denotes the power set. Consequently, we have $\mathcal{R} = \psi(\delta)$.}
Defining an exposure mapping function $f$ and the set of exposure conditions $\Delta$  is thus equivalent to identifying disjoint partitions of \([0, 1]^M\) — denoted as \(\mathcal{R}_1, \mathcal{R}_2, \ldots, \mathcal{R}_{|\Delta|}\), with the following properties:
\begin{enumerate}
    \item Each \(\mathcal{R}_k\) is a subset of \([0,1]^M\) and is non-empty, \(\mathcal{R}_k \neq \varnothing\);
    \item The union of all subsets covers \([0,1]^M\), \(\bigcup_{k=1}^{|\Delta|} \mathcal{R}_k = [0,1]^M\);
    \item The subsets are mutually exclusive, \(\mathcal{R}_{k} \cap \mathcal{R}_{k'} = \varnothing\) for all \( k \neq k' \).
\end{enumerate}

Accordingly, we define a set of exposure conditions \(\Delta = \{\delta_1, \delta_2, \ldots, \delta_{|\Delta|}\}\) such that \( \delta_k = h(\vr) \) for all \( \vr \in \mathcal{R}_k \). Therefore, to connect all notations, we have $f(\bm{z}, \theta_i)=h(g(\bm{z}, \theta_i))=h(\vr_i)=\delta_k$ and $\vr_i \in \mathcal{R}_k$.
With an exposure condition $\delta_k$ that corresponds to a subspace $\mathcal{R}_k \subset [0,1]^M$, the estimand of the average potential outcome $\mu^f(\delta_k)$ can be written as $\mu(\mathcal{R}_k) := \mu^f(\delta_k)$ with abuse of notation.

\begin{figure}
    \centering
    \includegraphics[width=\linewidth]{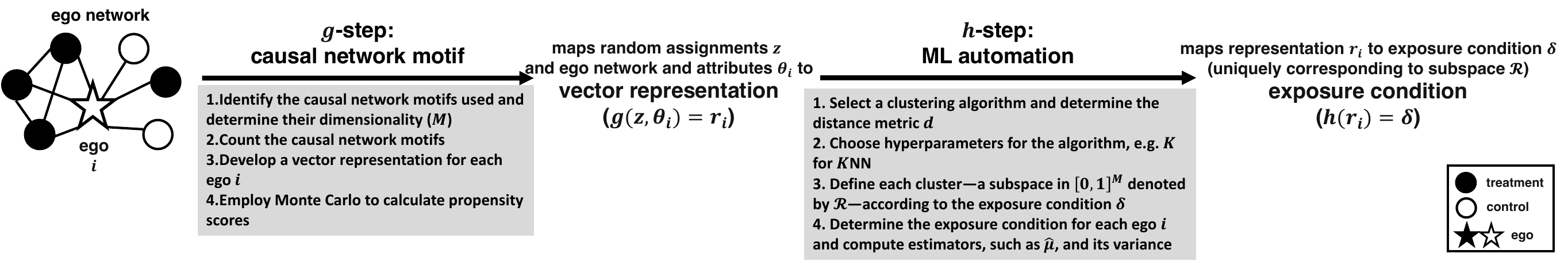}
    \caption{Diagram of our two-part approach compared to the conventional exposure mapping framework}
    \label{fig:diagram}
\end{figure}
We outline the pipeline of our two-part approach in Figure~\ref{fig:diagram}. Starting with the $N$-dimensional random assignment vector $\vZ$ (and its realization $\vz$), along with unit $i$'s network attributes and other covariates ($\theta_i$), we specify an $M$-dimensional random vector for the causal motif representation, denoted as $\vR_i$ (and its realization $\vr_i$). The support of $\vR_i$ is $[0, 1]^M$ (e.g., $\vr_i = (0, 0.12, \dots, 0.82)$). The exposure condition for unit $i$ is then uniquely determined based on which disjoint subspace in $[0,1]^M$ the vector representation $\vr_i$ falls into. Specifically, for each exposure condition $\delta \in \Delta$ and its corresponding subspace $\mathcal{R} \subset [0,1]^M$, we have $h(\vr_i)=\delta$  if $\vr_i \in \mathcal{R}$. The process of partitioning $[0,1]^M$ into disjoint subspaces ($\mathcal{R}_1, \mathcal{R}_2, \dots, \mathcal{R}_{|\Delta|}$) is accomplished through clustering algorithms. 

\section{Characterizing Interference by Causal Network Motif Representation}
\label{ref:motifs}

\subsection{Causal network motif }
The first step of our two-part approach is to specify the mapping $g$, which is to find an $M$-dimensional causal network motif representation $\vR_i$ for each unit $i$.
Network motifs~\citep{milo2002network} characterize the patterns of interactions among an ego's neighborhood. Network motifs provide a natural and interpretable way to characterize the local structure of interactions beyond just counts of friends or connections as is commonly done in the exposure mapping framework. 
We propose to use  \textit{causal network motifs}, which can characterize not only assignment conditions in the network neighborhood, but also ego network structure and individual attributes.\footnote{Although we do not explicitly illustrate it,  our approach is also adaptive to the neighbor node attributes and edge attributes. For example, a dimension can be fraction of treated female neighbors among all female neighbors, or fraction of treated strong ties (i.e., with interaction frequency greater than a cutoff) among all strong ties. }  
Accounting for network structure is a crucial missing component in the network interference literature regarding the specification of the exposure mapping function~\citep{ugander2013graph,aronow2017estimating}. According to network theories such as structural diversity~\citep{aral2011diversity,ugander2012structural} and the weak tie hypothesis~\citep{granovetter1973strength}, a unit's potential outcome may depend on the treatment assignments of their neighbors and their network structure. These dependencies can also be reflected by different causal network motifs.

\begin{figure}
    \centering
    \includegraphics[width=0.9\linewidth]{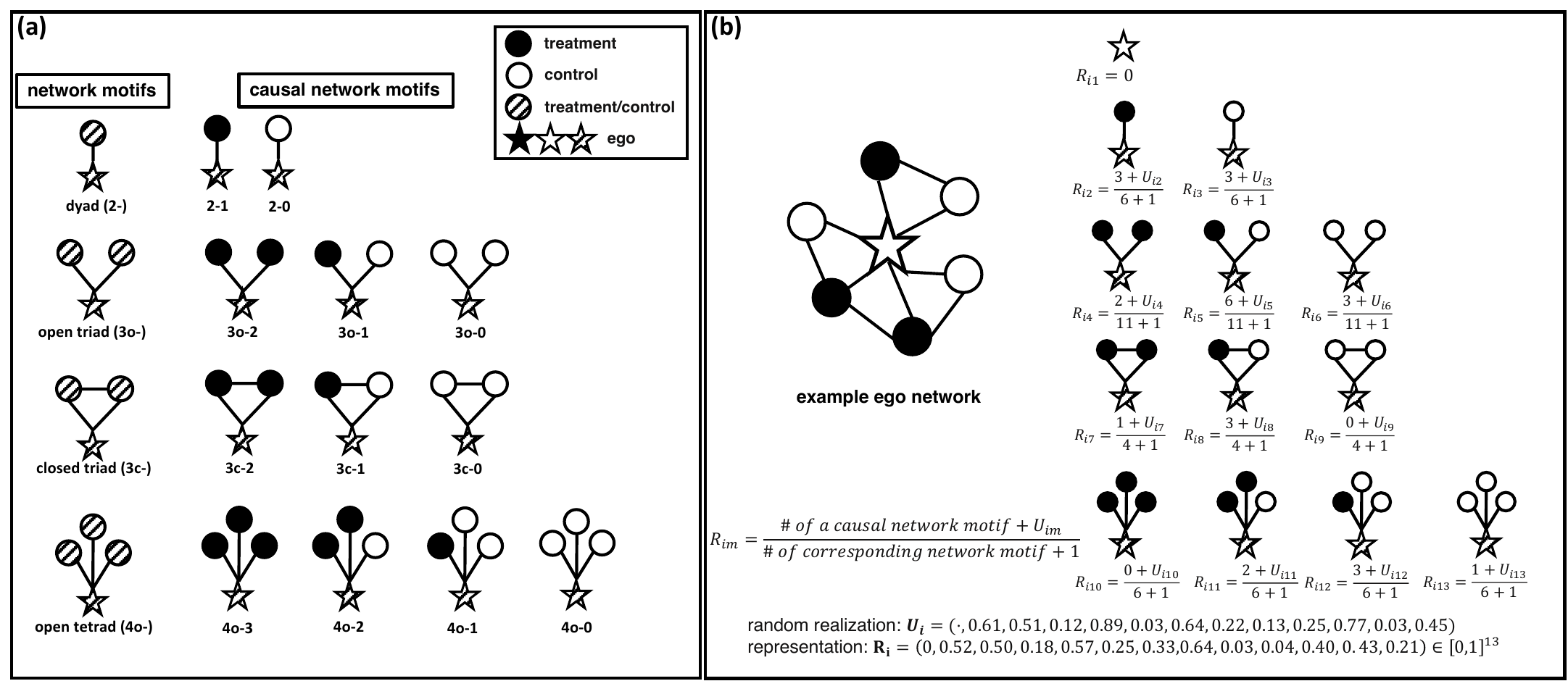}
    \caption{\scriptsize \raggedright Illustration of causal network motifs. \textbf{(a) Examples of causal network motifs.} Solid nodes indicate treatment, hollow nodes indicate control, and shaded nodes indicate that they could be in treatment or control. The star node for each network is the ego. The first pattern in each row represents conventional network motifs without assignment conditions, followed by corresponding causal network motifs. Our causal network motif representation is constructed by dividing the count of a causal network motif by the count of the corresponding network motif. The labels below each network motif indicate the naming: for example, an open triad where one neighbor is treated is named 3o-1 (the second causal network motif of the third row). \textbf{(b) Construction of causal network motif representation.} This illustration represents an ego network with treatment assignments and its corresponding causal network motif representation, considering $1$-hop network interference. The first dimension corresponds to the random assignment received by unit $i$ (denoted as $Z_{i1}$). Each subsequent dimension represents a causal network motif, calculated as the number of a specific causal network motif divided by the total number of corresponding network motifs. Random noise ($U_{im}$) is introduced to adjust the values so that each dimension maintains full support within the range $[0, 1]$.}
    \label{fig:causal_network_motif}
\end{figure}

Figure~\ref{fig:causal_network_motif}(a) illustrates our construction of causal network motif representation.\footnote{
It should be noted that although our current illustration concentrates on 1-hop ego networks, practitioners are able to utilize motifs that extend beyond 1-hop ego networks. This extension involves a trade-off between computational efficiency and estimation performance. Furthermore, our downstream machine learning approaches facilitate the automatic selection of significant motifs, even when dealing with high-dimensional causal network motif representations.
} 
The first dimension is the treatment assignment of unit $i$, i.e. $R_{i1} = Z_i$.
For the rest of dimensions ($m>1$), we should first specify the network motifs and then causal network motifs. 
 Each dimension ($m>1$) is the fraction of a causal network motif over the corresponding network motif.
\begin{equation*}
    R_{im} = \frac{\text{\# of a causal network motif}}{\text{\# of the corresponding network motif}} \text{ for } m>1.
\end{equation*}
This is the normalized number of causal network motifs over the total number of corresponding network motifs. For instance, it can be the ratio of fully treated closed triads (3c-2) to all closed triads (3c). In our running example, a higher proportion of treated closed triads suggests that a majority of a user’s strong ties—evidenced by a large number of mutual friends—are treated.

As previously discussed, selecting causal network motifs and determining the dimensionality \(M\) are akin to feature engineering in conventional machine learning. Practitioners should carefully consider both the availability and the relevance of features when integrating new ones. Similarly, including a more diverse array of causal network motifs generally enhances performance in estimation, mirroring the advantages of comprehensive feature inclusion in standard machine learning. However, it is essential to balance the trade-off between the computational demands of analyzing complex motifs and their marginal benefits to the estimation process. Our practical recommendation is to initially include as many causal network motifs as is computationally feasible, and subsequently refine a proper distance metric to emphasize more important dimensions (see Section~\ref{sec:distance}).

There are benefits for normalizing the causal network motif representation by the total number of corresponding network motifs. For example, if a partition is defined to include only units where the number of fully treated closed triads exceeds 10, then units with fewer than 10 open triads would inherently have a zero probability of inclusion in this partition.\footnote{Although our method of construction proves empirically useful, it is not the only feasible approach. For example, another possible construction is $R_{im} = \frac{1}{\text{\# of a causal network motif}}$. In addition, one dimension of $\vR_i$ might be more complex; for example, it could represent the fraction of disjoint components with at least one treated neighbor among all, reflecting structural diversity~\citep{ugander2012structural}.}

\subsection{Representation Invariance}
An essential concept for causal network representations is ``representation invariance.'' 

\begin{definition}[Representation Invariant to $\bm{z}$]
A mapping $g: \{0, 1\}^N \times \Theta \rightarrow [0,1]^M$ is representation invariant to the treatment assignment vector $\vz$ if 
$g(\vz, \theta_i) = g(\vz, \theta_j)$ for all $i \neq j$.
\end{definition}

This definition ensures that all units yield the same output from function $g$ (i.e., causal network motif representation) when the treatment assignment vector is  $\vz$.

\begin{remark}
The mapping $g$, as per our causal network motif construction, is representation invariant to $\bm{1}$ and $\bm{0}$. 
\end{remark}
We can thus define $\vr^{(1)}, \vr^{(0)} \in [0,1]^M$ such that $\vr^{(1)} = g(\bm{1},\theta_i)$ and $\vr^{(0)} = g(\bm{0},\theta_i)$ for all $i \in \mathcal{N}$, which enables estimation of the global average treatment effect.

\subsection{Adjusting Construction to Satisfy Positivity Requirement} 
Note that merely calculating the ratio does not guarantee full support in $[0, 1]$ for each dimension. To address this issue, we incorporate randomness in our computations as follows:
when calculating each $R_{im}$ (for $m > 1$), we add a random component --- $U_{im}$, which is an i.i.d. draw from the continuous uniform distribution $\text{Unif}([0, 1])$. This approach ensures that each $R_{im}$ for $m > 1$ has full support within $[0,1]$. Thus, the actual causal network motif representation is given by:
\begin{equation*}
    R_{im} = \frac{\text{\# of a causal network motif} + U_{im}}{\text{\# of corresponding network motifs} + 1} \quad \text{for } m > 1.
\end{equation*}
This technique also helps address the issue of having zero-valued denominators in the causal network motif representation, which can occur when the number of specific network motifs is zero.

Although representation invariance to $\bm{1}$ and $\bm{0}$ is not directly preserved, this issue can be effectively resolved by adjusting our interpretation of the function $g$. To restore representation invariance, we can allow $g$ to incorporate the randomness $\bm{U}_i$ as input, i.e., $\vR_i = g(\vZ, \bm{U}_i, \theta_i)$. We can appropriately set the values of $\bm{U}_i$ for all $i$: for $\vz=\bm{1}$, $\bm{U}_i$ is set to be $\vr^{(1)}$, i.e., $1$ for dimensions that count full treatment conditions (e.g., 2-1, 3c-2, and 3o-2) and to 0 for dimensions that count full control conditions (e.g., 2-0 and 3c-0). Likewise we set  $\bm{U}_i=\vr^{(0)}$ for $\vz=\bm{0}$. This approach ensures that $g(\bm{1}, \vr^{(1)}, \theta_i) = (\vr^{(1)}, \vr^{(1)})$ and $g(\bm{0}, \vr^{(0)}, \theta_i) = (\vr^{(0)}, \vr^{(0)})$ for all $i$, thereby maintaining representation invariance.

Figure~\ref{fig:causal_network_motif}(b) illustrates the calculation of each dimension's value for the example ego network after incorporating a random number $U_{im}$. The addition of $U_{im}$ subtly adjusts the ratio of causal network motifs to network motifs, ensuring that each dimension achieves full support within the interval $[0,1]$. For instance, the 6th dimension was initially calculated as 3/11 ($\approx$0.27), but after incorporating randomness, it adjusted to 0.25.

\section{Machine Learning to Specify Exposure Mapping}
\label{ref:algo}
As previously mentioned, the lack of automation for specification poses a significant challenge in the exposure mapping framework. To address this, we employ machine learning algorithms. Specifically, \(h\) employs an unsupervised learning (clustering) approach to partition the space \([0,1]^M\) into \(|\Delta|\) distinct subsets, denoted as \(\{\mathcal{R}_1, \mathcal{R}_2, \ldots, \mathcal{R}_{|\Delta|}\}\). This method effectively transforms the task into a clustering problem, a core approach in machine learning.

The performance of different clustering algorithms, which generate various sets of exposure conditions, can vary significantly, even within the same algorithm family. For example, the tree-based algorithm presented in the previous version~\citep{yuan2021causal} and nearest neighbors-based algorithms discussed in this section, may produce different exposure mappings due to variations in parameters. In this section, we first discuss the assumptions that justify the bias towards the global average treatment effect (\(\tau\)) in our estimates.\footnote{It is important to note that while the HT estimator is unbiased towards the estimand \(\tau^f\), \(\tau^f\) often does not equal \(\tau\) unless the exposure mapping \(f\) is precisely specified, which is highly challenging in practice.} Subsequently, we propose a clustering framework based on distance metrics and discuss one implementation using the \(K\)-nearest neighbors ($K$NN) algorithm.

\subsection{Monotonic Interference}
\label{sec:mono_inter}
Before introducing the process of specifying the mapping $h$, we discuss monotonic interference assumptions that are later used to assess estimators under different algorithms or their parameters.

\begin{assumption}[Monotonic interference]
\begin{enumerate}
\item[]
\item (Non-Negative Interference) We assume $y_i(\bm{z}) \geq y_i(\bm{z'})$ 
if $z_i \geq z_i'$ for $i=1,2,...,N$.
\item (Non-Positive Interference) We assume $y_i(\bm{z}) \leq y_i(\bm{z'})$ 
if $z_i \geq z_i'$ for $i=1,2,...,N$.
\end{enumerate}
\label{assumption:neg}
\end{assumption}
These two assumptions are realistic in many settings. Imagine that a group of people are targeted to adopt a new technology. People who are more likely to adopt it when more of their friends have already adopted it. This effect may spread to friends' friends or ultimately everyone else in the world. 
In this setting, the non-negative interference assumption is satisfied if $Y_i$ represents whether $i$ adopts this new technology, which means others' adoption would either increase $i$'s likelihood of adoption or at least would not decrease it. 
Similarly, when the intervention is to receive a vaccine and $Y_i$ is a binary variable representing the infection of the corresponding disease, the non-positive interference assumption is likely satisfied. 
Similar assumptions have been made in many previous related works, such as \cite{pouget2018optimizing} and \cite{aronow2021spillover}.\footnote{Note that there are also instances where monotonic interference is not applicable. For instance, in the context of fashion adoption, an individual's likelihood of adopting a fashion product may increase if they observe people they admire adopting it, yet decrease if they see it adopted by those they dislike. In practice, it is suggested that practitioners use their prior knowledge or common sense to justify that monotonic interference is reasonable in their specific settings. }

Despite the challenge of correctly specifying the exposure mapping function $f$ \citep{savje2023causal}, a misspecified exposure mapping may still be informative under assumptions of monotonic interference.
\begin{proposition}
\begin{enumerate}
    \item If $f$ is correctly specified, $\Exp_{\vZ} [\hat\tau_{\text{HT}}^f] = \tau^f = \tau$; 
    \item Under the non-negative interference, if $f$ is misspecified, then $\Exp_{\vZ} [\hat\tau_{\text{HT}}^f] = \tau^f   \leq \tau $;
    \item Under the non-positive interference, if $f$ is misspecified, then $\Exp_{\vZ} [\hat\tau_{\text{HT}}^f] = \tau^f  \geq \tau $.
\end{enumerate}
\label{prop:misspecify}
\end{proposition}

\prf{See \textit{Appendix~\ref{proof:misspecify}.}}

The implication of Proposition \ref{prop:misspecify} is as follows: when \( f \) is correctly specified, the HT estimator unbiasedly measures both the estimand (condition expectation) \( \tau^f \) and the global average treatment effect \( \tau \). When \( f \) is misspecified—a practically common scenario—the  monotonic interference assumption may guide the selection among various estimators. For instance, under non-negative interference, the largest HT estimator introduces the least bias in expectation, but under non-positive interference the smallest estimator does.\footnote{These insights also extend to the Hájek estimator, which is approximately unbiased though its bias is more complex to analyze \citep{eckles2016design,khan2021adaptive}.}

\subsection{Clustering and Distance Metrics}
\label{sec:distance}

We explore the use of clustering algorithms to specify exposure conditions based on causal network motifs. Most clustering algorithms employ a distance metric. Define \( d \) as a distance function applicable to pairs of causal motif representations within the space \([0,1]^M\), where \( d: [0,1]^M \times [0,1]^M \rightarrow \mathbb{R}_+ \). The goal is to optimize a specific criterion, such as minimizing the total distances of units within the same clusters.
\cite{yuan2021causal} falls under this framework and describes a tree-based clustering algorithm that partitions \([0,1]^M\) into disjoint subspaces (exposure conditions).\footnote{It can be understood as \( d(\vr, \vr') = 0 \) if both \(\vr\) and \(\vr'\) belong to the same leaf in the tree, and \( d(\vr, \vr') = 1 \) otherwise.} In this paper, we focus on a set of norm-based distance metrics:

\begin{itemize}
    \item \textit{Identical weights}. Each dimension is treated with equal weight 
        \begin{equation*}
        d_\text{I}(\vr, \vr') = ||\vr - \vr'||_1.
        \end{equation*}
    Here, $||\bm{x}||_1$ denotes the $\ell_1$ norm, measuring the Manhattan distance between vectors.

    \item \textit{Regression coefficients}. Let $\boldsymbol{\beta}$ be an $M$-dimensional coefficient vector, then define:
        \begin{equation}
        d_\text{C}(\vr, \vr') = || \boldsymbol{\beta} \circ (\vr - \vr') ||_1.
        \label{eq:reg:coef}
        \end{equation}
    Here $\circ$ represents the element-wise product. Intuitively, this approach scales up the dimensions of the causal motif representation that are more predictive of the outcome, implying that even small discrepancies between $\vr$ and $\vr'$ in these critical dimensions significantly impact the distance. Inspired by \cite{imai2013estimating}, a distance metric in a causal inference task may consider both treatment variables (causal network motifs denoted by $\bm{R}_i$) and covariates (demographics and network motifs, denoted by $\bm{X}_i$). Thus we consider the following regression-based metrics:
\begin{enumerate}
    \item \textit{regression}: Regress \(Y_i\) on \(\vR_i\) and let \(\boldsymbol{\beta}\) in Eq.~\eqref{eq:reg:coef} be the regression coefficients of \(\vR_i\). 
    \item \textit{regression\_cov}: Regress \(Y_i\) on the causal network motif representation \(\vR_i\) along with covariates \(\bm{X}_i\), utilizing only the regression coefficients of \(\vR_i\) for \(\boldsymbol{\beta}\).
    \item \textit{lasso\_cv}: Employ Lasso regression (with cross-validation for hyperparameter tuning) to regress \(Y_i\) on \(\vR_i\), and let \(\boldsymbol{\beta}\) be the regression coefficients of \(\vR_i\).
    \item \textit{lasso\_cv\_cov}: Employ Lasso regression (with cross-validation for hyperparameter tuning) to regress \(Y_i\) on \(\vR_i\) and \(\bm{X}_i\), and let \(\boldsymbol{\beta}\) be the regression coefficients of \(\vR_i\).
\end{enumerate}
\end{itemize}

We next introduce the concept of a \textit{properly specified distance metric}:
\begin{definition}[Properly Specified Distance Metric]
For all $\vz, \vz' \in \{0,1\}^N$, $\vr_i=g(\vz, \theta_i)$ and $\vr_i'=g(\vz', \theta_i)$ for all $i \in \mathcal{N}$:
\begin{enumerate}
\item Under the non-negative interference assumption, a distance metric $d$ is properly specified if, for all $i \in \mathcal{N}$, $d(\bm{r}^{(1)}, \vr_i) \geq d(\bm{r}^{(1)}, \vr_i')$ implies $y_i(\vz) \leq y_i(\vz')$, and $d(\bm{r}^{(0)}, \vr_i) \geq d(\bm{r}^{(0)}, \vr_i')$ implies $y_i(\vz) \geq y_i(\vz')$.
\item Under the non-positive interference assumption, a distance metric $d$ is properly specified if, for all $i \in \mathcal{N}$, $d(\bm{r}^{(1)}, \vr_i) \geq d(\bm{r}^{(1)}, \vr_i')$ implies $y_i(\vz) \geq y_i(\vz')$, and $d(\bm{r}^{(0)}, \vr_i) \geq d(\bm{r}^{(0)}, \vr_i')$ implies $y_i(\vz) \leq y_i(\vz')$.
\end{enumerate}
\end{definition}
Intuitively, a properly specified distance metric should properly reflect the relationship between the potential outcome \(y_i\) and the distance between causal network motif representations \(\vr_i\) and \(\vr^{(1)}\) (or \(\vr^{(0)}\)). When \(\vr_i\) is farther from \(\vr^{(1)}\) (or \(\vr^{(0)}\)) according to the distance metric \(d\), the potential outcome \(y_i\) should change monotonically.

Determining the appropriateness of a distance metric is essentially an empirical task, often addressed using cross-validation which is common in standard machine learning. In practice, incorporating a wide set of causal network motifs followed by reasonable distance metric learning algorithms and proper validation, such as Lasso with hyperparameter cross-validation, can lead to effective distance metrics. Conversely, a poorly defined distance metric might lead to inferior estimations.

\subsection{$K$ Nearest Neighbors Specification}
We propose a nearest neighbors approach to specify the exposure mapping, primarily aiming to estimate the global average treatment effect, such as the impact of universally granting new messaging feature access. This method contrasts with the tree-based algorithm described in \cite{yuan2021causal}, which is designed to specify all exposure conditions and estimate average potential outcomes. For global average treatment effects, it suffices to specify only $\delta^{(1)}$ and $\delta^{(0)}$, which represent the two exposure conditions that approximate the fully treated and non-treated scenarios, respectively.

This nearest neighbors approach also requires that the mapping $g$ be representation invariant to $\vz=\bm{1}$ and $\vz=\bm{0}$.\footnote{The tree-based algorithm \cite{yuan2021causal} similarly requires representation invariance to $\vz=\bm{1}$ and $\vz=\bm{0}$.} We let $\vr^{(1)} = g(\bm{1}, \theta_i)$ and $\vr^{(0)} = g(\bm{0}, \theta_i)$ for all $i \in \mathcal{N}$. Our construction of causal network motifs described in Section~\ref{ref:motifs}  satisfies this property.

In the $K$ nearest neighbor ($K$NN) specification, determining the number $K$ is crucial as it affects how we approximate the fully treated or untreated scenarios (to $\vr^{(1)}$ or $\vr^{(0)}$). Since a distance metric \( d \) and the number \( K \) uniquely define an exposure mapping, we can superscript \( h \) by \( h^{d,K} \). Consequently, we can rewrite \( \mu^f \) as \( \mu^{d,K} \) and \( \tau^f \) as \( \tau^{d,K} \) for the nearest neighbors specification.

To estimate the global average treatment effect, our focus should be solely on the two exposure conditions corresponding to the fully treated and non-treated scenarios. Therefore, we can define the set of exposure conditions: $\Delta = \{\delta^{(1)}, \delta^{(0)}, \delta^{\text{otw}}\}$:

\[ h^{d,K}(\vR) = \begin{cases}
      \delta^{(1)} & \quad \text{ if } \left| \{i \in \mathcal{N} | d(\vr_i, \vr^{(1)}) \leq  d(\vR, \vr^{(1)}) \} \right| \leq K \\
      \delta^{(0)} & \quad \text{ if } \left| \{i \in \mathcal{N} | d(\vr_i, \vr^{(0)}) \leq  d(\vR, \vr^{(0)}) \} \right| \leq K \\
      \delta^{\text{otw}} & \quad \text{otherwise}
   \end{cases}\]

As each exposure condition corresponds to a subspace, we can define subspaces $\mathcal{R}^{(1)}$ and $\mathcal{R}^{(0)}$ such that $\mathcal{R}^{(1)} = \{\vr \in [0,1]^M \mid h(\vr) = \delta^{(1)}\}$ and $\mathcal{R}^{(0)}  = \{\vr \in [0,1]^M \mid h(\vr) = \delta^{(0)}\}$ to compute the global average treatment effect:\footnote{If the primary objective is to estimate the average potential outcome given a specific representation $\vr$, we can create an exposure condition that is ``close enough'' to $\vr$. This approach allows us to answer questions such as ``What would be the average potential outcome for units who had half of their neighbors treated?''}
\begin{equation*}\vspace{-.3cm}
\hat{\tau}^{d,K} = \hat{\mu}^{d,K}(\delta^{(1)}) -  \hat{\mu}^{d,K}(\delta^{(0)}) = \hat{\mu}\left(\mathcal{R}^{(1)}\right) - \hat{\mu}\left(\mathcal{R}^{(0)}\right).
\end{equation*}

\noindent\textbf{Honest splitting.} Similar to the honest splitting proposed in~\cite{athey2016recursive}, we randomly and equally split all units ($\mathcal{N}$) into a distance metric set $\mathcal{N}^{\text{dm}}$ and estimation set $\mathcal{N}^{\text{est}}$. We use $\mathcal{N}^{\text{dm}}$  to learn regression coefficients and construct the distance metric $d$ and then use $\mathcal{N}^{\text{est}}$ only to perform estimation and inference.

\subsection{Finding Optimal $K$}

From Proposition~\ref{prop:misspecify}, we understand that under the assumption of non-negative (non-positive) interference, the largest (smallest) estimator should be the least biased. Thus, we can explore a set of potential values for \( K \) that satisfy the positivity requirements for \( \delta^{(1)} \) and \( \delta^{(0)} \), denoted by \( \mathbb{K} \). For instance, under the non-negative interference assumption, we can construct the following estimator:
$\hat\tau^* = \text{max}_{K \in \mathbb{K}} \left( \hat{\mu}^{d,K}(\delta^{(1)}) - \hat{\mu}^{d,K}(\delta^{(0)}) \right)$ and we have $\hat\tau^* \leq \tau$.
This implies that the \( K \) leading to the largest estimator while still satisfying the positivity requirements yields the least biased estimator. 

Given the practical challenge in correctly specifying an exposure mapping function, obtaining an unbiased estimator of the global average treatment effect in scenarios with interference is practically unlikely. Consequently, we adopt a pragmatic approach, opting for estimators that are biased but are the least biased among the available options.

Under monotonic interference assumptions, we can also observe a monotonic trend between \( K \) and the bias in estimating global average treatment effects:
\begin{proposition}
\begin{enumerate}
\item Under the assumption of non-negative interference and a properly specified distance metric $d$, as $K$ increases (within a range that maintains positivity), the expected value of $\hat{\tau}_{\text{HT}}^{d,K}$ decreases, and the bias of $\hat{\tau}_{\text{HT}}^{d,K}$ relative to $\tau$ (i.e., $|\hat{\tau}_{\text{HT}}^{d,K} - \tau| $) increases.
\item Under the assumption of non-positive interference and a properly specified distance metric $d$, as $K$ increases (within a range that maintains positivity), the expected value of $\hat{\tau}_{\text{HT}}^{d,K}$ increases, and the bias of $\hat{\tau}_{\text{HT}}^{d,K}$ relative to $\tau$ (i.e., $|\hat{\tau}_{\text{HT}}^{d,K} - \tau| $) also increases.
\end{enumerate}
\label{prop:K}
\end{proposition}

\prf{
See \textit{Appendix~\ref{appendix:K}.}
}

Intuitively, this proposition suggests that by  utilizing a narrower region around \( \vr^{(1)} \) (or \( \vr^{(0)} \)) to approximate the fully treated or non-treated scenarios (\( \mathcal{R}^{(1)} \) and \( \mathcal{R}^{(0)} \), respectively), obtained via a smaller \( K \), we can achieve a less biased estimate of the true global average treatment effect.

Empirically, under the monotonic interference assumption, selecting a smaller value of \( K \) for the \( K \)NN algorithm may be advantageous, provided that the positivity requirement is met and the variance remains within a preset bound. Under a preset variance upper bound \( \bar{V} \), the optimal \( K^* \) can be defined as \( K^* = \{ K \in \mathbb{K} \mid \text{Var}(\widehat{\tau}^{d,K}) \leq \bar{V} \} \).
This strategy helps yield estimators with small bias relative to  global average treatment effects while maintaining reasonably small variances.\footnote{Note that a monotonic trend between $K$ and variance is not inherently guaranteed, although our simulations show that tends to decrease in \(K\).}

\subsection{Variance Estimation}
Finally, we present the variance estimations. Since our approach falls within the exposure mapping framework, all the variance estimation methods in \cite{aronow2017estimating} are applicable. Note that \cite{aronow2017estimating} requires a correctly specified exposure mapping. Although our framework does not require this, their conclusion still applies for any subspace \( \mathcal{R}_k \in [0,1]^M\), which may uniquely define an exposure condition. We briefly summarize the variance estimation as follows.
\begin{align*}
\text{Var}(\hat{\mu}_{\text{HT}}(\mathcal{R}_k)) = & \sum_{i \in \mathcal{N}} \mathbb{P}[\vR_i \in \mathcal{R}_k] \left( 1 - \mathbb{P}[\vR_i \in \mathcal{R}_k] \right) \left[ \frac{\mu_i(\mathcal{R}_k)}{\mathbb{P}[\vR_i \in \mathcal{R}_k]} \right]^2 \\
& + \sum_{i, j \in \mathcal{N}, \, i \neq j} \left( \mathbb{P}[\vR_i, \vR_j \in \mathcal{R}_k] - \mathbb{P}[\vR_i \in \mathcal{R}_k] \mathbb{P}[\vR_j \in \mathcal{R}_k] \right) \left( \frac{\mu_i(\mathcal{R}_k) \mu_j(\mathcal{R}_k)}{\mathbb{P}[\vR_i \in \mathcal{R}_k] \mathbb{P}[\vR_j \in \mathcal{R}_k]} \right)
\end{align*}

The variance is estimated as 
\begin{align}
\widehat{\text{Var}}(\hat{\mu}_{\text{HT}}(\mathcal{R}_k )) = & \sum_{i: \vR_i \in \mathcal{R}_k} \left(1-\mathbb{P}[\vR_i \in \mathcal{R}_k] \right) \left( \frac{Y_i}{\mathbb{P}[\vR_i \in \mathcal{R}_k]} \right)^2 \\
& + \sum_{i\neq j: \vR_i, \vR_j \in \mathcal{R}_k\text{ and }\mathbb{P}[\vR_i, \vR_j \in \mathcal{R}_k ]\neq0} \frac{ \mathbb{P}[\vR_i, \vR_j \in \mathcal{R}_k] - \mathbb{P}[\vR_i \in \mathcal{R}_k] \mathbb{P}[\vR_j \in \mathcal{R}_k] }{ \mathbb{P}[ \vR_i, \vR_j \in \mathcal{R}_k ]} \frac{Y_i Y_j}{\mathbb{P}[\vR_i \in \mathcal{R}_k] \mathbb{P}[\vR_j \in \mathcal{R}_k]} \label{eq:varest}\\ 
& + \sum_{i\neq j: \vR_i, \vR_j \in \mathcal{R}_k\text{ and }\mathbb{P}[\vR_i, \vR_j \in \mathcal{R}_k ]=0} \frac{1}{2} \left( \frac{Y_i^2}{\mathbb{P}[\vR_i\in \mathcal{R}_k ]} + \frac{Y_j^2}{ \mathbb{P}[\vR_j \in \mathcal{R}_k ]}\right) 
\end{align}

The variance for the estimator of global average treatment effects $\hat{\tau}^{d,K}$  is
\begin{equation*}
{\text{Var}}(\hat{\tau}_{\text{HT}}^{d,K}) = {\text{Var}}(\hat{\mu}_{\text{HT}}(\mathcal{R}^{(1)})) + {\text{Var}}(\hat{\mu}_{\text{HT}}(\mathcal{R}^{(0)})) - 2 \text{Cov}\left[\hat{\mu}_{\text{HT}}(\mathcal{R}^{(1)}), \hat{\mu}_{\text{HT}}(\mathcal{R}^{(0)})\right],
\end{equation*}
and its variance estimator is 
\begin{equation*}
\widehat{\text{Var}}(\hat{\tau}_{\text{HT}}^{d,K}) = \widehat{\text{Var}}(\hat{\mu}_{\text{HT}}(\mathcal{R}^{(1)})) + \widehat{\text{Var}}(\hat{\mu}_{\text{HT}}(\mathcal{R}^{(0)})) - 2 \widehat{\text{Cov}}\left[\hat{\mu}_{\text{HT}}(\mathcal{R}^{(1)}), \hat{\mu}_{\text{HT}}(\mathcal{R}^{(0)})\right].
\end{equation*}

Note that Eq.~\eqref{eq:varest} contains $\mathbb{P}[\vR_i, \vR_j \in \mathcal{R}_k] - \mathbb{P}[\vR_i \in \mathcal{R}_k] \mathbb{P}[\vR_j \in \mathcal{R}_k]$, which equals zero if $\vR_i \indep \vR_j$. We discuss, under Bernoulli or cluster randomization, how to account for this pairwise dependency, and how computational efficiency for variance estimation can be improved by focusing solely on non-independent pairs. Following \cite{aronow2017estimating}, \textit{Appendix~\ref{sec:theoretical}} also provides detailed discussions on covariance estimation, demonstrates that the estimators employed are conservative, and explores asymptotic properties including consistency and asymptotic properties for confidence intervals.

\noindent\textbf{Test of existence of network interference.}
With the variance estimators, we can also perform a statistical test on the existence of interference.
Specifically, we define the two subspaces: (1) $\mathcal{R}^{(1)}$ determined by the distance metric \( d \) and value of \( K \);
and (2) $\mathcal{R}^{\text{treat}}$ defined as $\{ \vr \in [0,1]^M | \vr_1=1\}$.

Define the test:
\vspace{-0.3cm}\[T = \frac{\hat{\mu}(\mathcal{R}^{(1)}) - \hat{\mu}(\mathcal{R}^{\text{treat}})}{ 
\sqrt{\widehat{\text{Var}}\left(\hat{\mu}(\mathcal{R}^{(1)})\right) + \widehat{\text{Var}}(\hat{\mu}(\mathcal{R}^{\text{treat}}) - 
2\widehat{\text{Cov}}(\hat{\mu}(\mathcal{R}^{(1)}), 
\hat{\mu}(\mathcal{R}^{\text{treat}}))}}\]

As $n \rightarrow \infty$, $T \sim N(0,1)$. The p-value can be calculated as
\( p\text{-value} = 2 \times (1 - \Phi(|T|)) \)
where $\Phi$ is the CDF  of the standard normal distribution. This test assesses the existence of network interference for the treatment group by comparing an exposure condition that approximates the fully treated scenario to the average of all treated samples. Note a similar test can be defined for the control group.

\section{Synthetic Experiments}

\subsection{Network Setup}
Since real-world data cannot establish definitive outcomes for all counterfactual scenarios, we start by employing a synthetic network to generate potential outcomes.
The Watts-Strogatz (WS) network is a classical network model that captures important real-world network properties including clustering and the ``small-world phenomenon''~\citep{watts1998collective}. This network model also provides more variety in network structure compared to other models, such as Erd\"os–R\'enyi model and stochastic block models.
One important parameter in the WS model is the rewiring rate which ranges in $[0,1]$; it controls the randomness  and the extent of clustering of the network. We chose a rewiring rate of $0.5$ which results in a diverse set of causal network motifs and helps illustrate the advantage of our approach compared to not accounting for motifs.\footnote{Watts-Strogatz network has also been widely used in previous work on network interference, including \cite{eckles2016design} and \cite{chin2019regression}.}

We evaluate two randomization approaches: Bernoulli randomization and graph cluster randomization~\citep{ugander2013graph,eckles2016design}. In Bernoulli randomization, for each $i \in \mathcal{N}$, we randomly assign the treatment $Z_i \sim \text{Bern}(0.5)$. In the case of graph cluster randomization, we apply the standard Kernighan–Lin graph clustering algorithm~\citep{kernighan1970efficient} recursively, yielding a total of 512 balanced (same-sized) graph clusters. We apply $\text{Bern}(0.5)$ on the cluster level.

We evaluate the following three sets of causal network motifs, respectively:
\begin{enumerate}

\item \textbf{Square-full}: As an illustration, we select various network motifs, including the proportion of treated neighbors (2-1), non-treated open triads (3o-0), treated open triads (3o-2), non-treated closed triads (3c-0), treated closed triads (3c-2), treated open tetrads (4o-3), and non-treated open tetrads (4o-0), as depicted in Figure~\ref{fig:causal_network_motif}. Furthermore, we distinguish between the proportions of treated neighbors with \( X_j=1 \), denoted as 2-1(1), and the proportions of treated neighbors with \( X_j=0 \), denoted as 2-1(0). This distinction helps account for heterogeneous interference effects related to the attributes of \( j \).\footnote{Relatedly, we also introduce \textbf{Square-nocov}, which employs the same set of causal network motifs as \textbf{Square-full} but excludes the two covariate-related dimensions.}
\item \textbf{Triad}: We exclude tetrad-related dimensions, as well as covariate-related dimensions, i.e., 2-1(1) and 2-1(0), from \textbf{Square-full}.
\item \textbf{\(q\)-frac} (benchmark): This configuration aligns with the fractional \(q\)  neighborhood exposure described by \cite{ugander2013graph}. It includes only two dimensions: \(Z_i\) and the fraction of treated neighbors (2-1).
\end{enumerate}

In the remainder of this section, we use different potential outcomes models and apply our approach to each of them to assess its performance under various scenarios.

\begin{figure}
    \centering
    \includegraphics[width=0.8\linewidth]{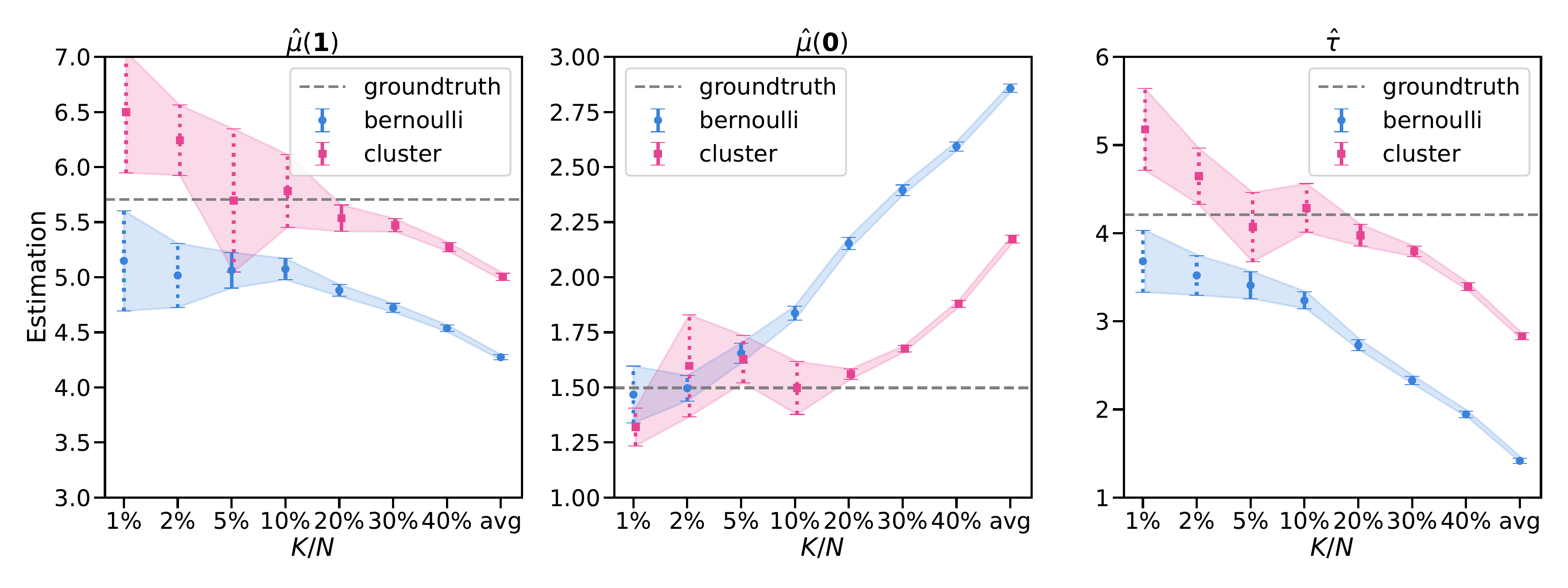}
    \includegraphics[width=0.8\linewidth]{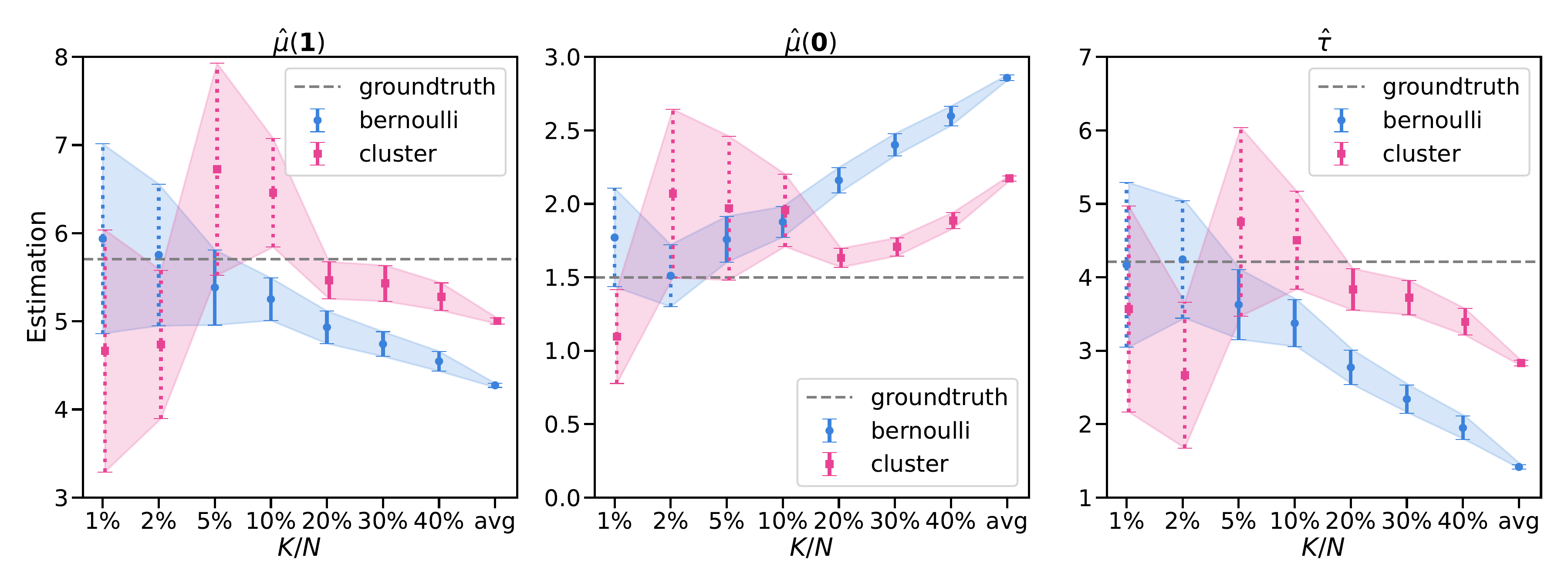}
    \caption{\raggedright \scriptsize
Results for the WS synthetic experiment with the baseline potential outcomes: (Upper) H\'ajek and (Lower) HT estimators. Blue and pink error bars represent Bernoulli and graph cluster randomization, respectively. From left to right, the panels illustrate the estimators for the counterfactual world where every unit is treated ($\mu(\mathbf{1})$), non-treated ($\mu(\mathbf{0})$), and the global average treatment effects ($\tau$). The $x$-label $K/N$ represents the fraction of the number of nodes used in the nearest neighbor exposure condition relative to the entire population. Error bars depict standard errors, wherein the dotted error bars indicate estimates that do not meet the positivity requirement. The dashed gray lines represent the ground truth. The label ``avg" denotes the approach of calculating average outcomes for treatment and/or control groups.}
    \label{fig:main}
\end{figure}

\subsection{Base Potential Outcome Model}

We first generated the potential outcomes for each node using the following function:
\begin{equation}
y_i(\vZ) = (1 + X_i) (1 + Z_i + \sum_{j \in \mathcal{N}_i} w_{ij} Z_j) + \varepsilon_i.\label{eq:base}
\end{equation}
Here, $w_{ij}$ is the weight for each edge: 
we first compute the number of common friends between $i$ and $j$, denoted by $\text{cf}_{ij}$;
then we compute $w_{ij} = \frac{ (X_j+1)\text{cf}_{ij}}{\sum_{j \in \mathcal{N}_i} (X_j+1) \text{cf}_{ij}} $.
This implies that neighbors sharing more mutual friends exert stronger interference. Pairs without mutual friends are presumed not to generate any interference.\footnote{Note that some nodes might lack neighbors with shared friends; in such instances, we assume that these nodes' outcomes remain unaffected by interference.}
This further underscores the potential significance of closed network motifs (like closed triads or squares) in characterizing interference.
$X_i$ is a binary covariate that interacts with the treatment assignments of both the ego nodes and other nodes. 
Here we assume that there are two types of nodes (e.g. different genders or other demographic groups) and that $X_i \sim \text{Bern}(0.5)$. The effect size and interference are in general larger when $X_i=1$, and the Gaussian noise is $\varepsilon_i\sim N(0,\frac{1}{16})$.
The inclusion of $X_j$ denotes heterogeneous interference based on neighbors' attributes. In other words, $X_j=1$ generally induces stronger interference spillover onto their neighbors compared to $X_j=0$. 
Note that this potential outcome adheres to the non-negative interference assumption.
In this synthetic data, when $Z_i=1$ for all $i\in \mathcal{N}$, the average potential outcome (or $\mu(\bm{1})$) equals $5.7$
when $Z_i=0$ for all $i\in \mathcal{N}$, the average potential outcome (or $\mu(\bm{0})$) equals $1.5$; therefore, the true global average treatment effect is $\tau = \mu(\bm{1}) - \mu(\bm{0}) = 4.2$.\footnote{Tree-based model results are presented in \textit{Appendix~\ref{sec:tree:res}}.}

We examine the nearest neighbors approach under different choices of $K$ for the evaluation of the average potential outcomes for the fully treated ($\mu(\bf{1})$), fully non-treated ($\mu(\bf{0})$) scenarios, and then estimate the global average treatment ($\tau$), in Figure~\ref{fig:main}. 
We apply our approach to both Bernoulli randomization and graph cluster randomization. As an illustration, we employ the regression distance metric in Figure~\ref{fig:main}.
Note that if $K$ is too small and fails to satisfy the positivity requirement, indicated by dashed error bars, estimations should be disregarded.
Our observations from the Bernoulli randomization reveal an increasing bias with $K$, denoted by a decreasing trend in the estimations of $\mu(\bf{1})$ and $\tau$, and an increasing trend for $\mu(\bf{0})$. This observation aligns with the conclusions from Proposition~\ref{prop:K}. At the same time, an inverse relationship is observed between the increasing $K$ and decreasing variance, reflecting an interesting bias-variance trade-off.

We also apply the nearest neighbors approach to graph cluster randomization. The bias-variance trade-off also applies to cluster randomization: When the positivity requirement is met,\footnote{Note that certain units are less likely to belong to two specific exposure conditions, $\mathcal{R}^{(1)}$ and $\mathcal{R}^{(0)}$, due to their clustering with closely embedded neighbors. Such nodes often display characteristics similar to feature 3c-2, tending towards values closer to either 1 or 0.
} an increase in $K$ results in a decrease in variance, but also an increase in bias. Interestingly, merging our approach and cluster randomization can enhance the precision of the estimation beyond what our approach achieves alone. For instance, $K=\lfloor 20\%N \rfloor$ is the smallest $K$ that satisfies positivity, and it exhibits less bias than the least biased estimation in the Bernoulli case. 
The results suggest that our approach can be effectively incorporated with graph cluster randomization.

We numerically present  coverage rates of confidence intervals in Table~\ref{tab:coverage}. We re-randomized the treatments and regenerated the potential outcomes and repeated the estimations 250 times to derive the numbers. We present the coverage rates of confidence intervals relative to the true global average treatment effects.\footnote{Note that under misspecification, these estimators are expected to be biased towards the ground truth; therefore, we do not expect all empirical coverage rates to exceed 90\%, 95\%, or 99\%.}

\begin{table}[t!]
\centering
\caption{ Coverage Rates of Confidence Intervals}\scriptsize
\label{tab:coverage}
\begin{threeparttable}
\begin{tabular}{@{}c|cccccc|ccccccc@{}}
\toprule
\textbf{K/N} & \begin{tabular}[c]{@{}l@{}}\textbf{HJ}\\ \textbf{90\%}\end{tabular} & \begin{tabular}[c]{@{}l@{}}\textbf{HT}\\ \textbf{90\%}\end{tabular} & \begin{tabular}[c]{@{}l@{}}\textbf{HJ}\\ \textbf{95\%}\end{tabular} & \begin{tabular}[c]{@{}l@{}}\textbf{HT}\\ \textbf{95\%}\end{tabular} & \begin{tabular}[c]{@{}l@{}}\textbf{HJ}\\ \textbf{99\%}\end{tabular} & \begin{tabular}[c]{@{}l@{}}\textbf{HT}\\ \textbf{99\%}\end{tabular} & \begin{tabular}[c]{@{}l@{}}\textbf{HJ}\\ \textbf{90\%}\end{tabular} & \begin{tabular}[c]{@{}l@{}}\textbf{HT}\\ \textbf{90\%}\end{tabular} & \begin{tabular}[c]{@{}l@{}}\textbf{HJ}\\ \textbf{95\%}\end{tabular} & \begin{tabular}[c]{@{}l@{}}\textbf{HT}\\ \textbf{95\%}\end{tabular} & \begin{tabular}[c]{@{}l@{}}\textbf{HJ}\\ \textbf{99\%}\end{tabular} & \begin{tabular}[c]{@{}l@{}}\textbf{HT}\\ \textbf{99\%}\end{tabular} \\ \hline
& \multicolumn{6}{c|}{\textit{Bernoulli Randomization}} & \multicolumn{6}{c}{\textit{Cluster Randomization}} \\
\textbf{0.01} & 0.86 & 1.00 & 0.91 & 1.00 & 0.98 & 1.00 & 0.78 & 0.92 & 0.84 & 0.96 & 0.92 & 0.98 \\
\textbf{0.02} & 0.84 & 1.00 & 0.89 & 1.00 & 0.96 & 1.00 & 0.76 & 0.94 & 0.85 & 0.97 & 0.94 & 0.99 \\
\textbf{0.05} & 0.08 & 0.99 & 0.14 & 1.00 & 0.36 & 1.00 & 0.80 & 0.81 & 0.87 & 0.94 & 0.94 & 1.00 \\
\textbf{0.1} & 0.00 & 0.03 & 0.00 & 0.14 & 0.00 & 0.76 & 0.90 & 0.58 & 0.96 & 0.75 & 0.99 & 0.96 \\
\textbf{0.2} & 0.00 & 0.00 & 0.00 & 0.00 & 0.00 & 0.00 & 0.57 & 0.68 & 0.67 & 0.81 & 0.86 & 0.96 \\
\textbf{0.3} & 0.00 & 0.00 & 0.00 & 0.00 & 0.00 & 0.00 & 0.00 & 1.00 & 0.00 & 1.00 & 0.00 & 1.00 \\
\textbf{0.4} & 0.00 & 0.00 & 0.00 & 0.00 & 0.00 & 0.00 & 0.00 & 0.06 & 0.00 & 0.37 & 0.00 & 0.98 \\
\bottomrule
\end{tabular}
\begin{tablenotes}
\tiny
\item \textit{Note: HJ = Hajek estimator, HT = HT estimator.}
\end{tablenotes}
\end{threeparttable}
\end{table}

We also examine the impacts of different causal network motif sets and distance metrics in Figure~\ref{fig:feature}. Analyzing various causal network motif sets, we find that $q$-frac (the benchmark representing fractional $q$ exposure mapping), which does not account for network structure beyond the fraction of treated neighbors, performs less well compared to the other two motif sets that incorporate more complex network structures. This difference is particularly evident when $K$ is small. This suggests the benefits of complex causal network motifs. 
The distinctions between the `Square-full' and `Triad' motif sets are less evident. However, at $K/N=5\%$, where none of the estimators violates the positivity requirement, the estimators from `Square-full' are slightly less biased than those from `Triad'.

Regarding distance metrics, for both `Square-full' and `Triad', the identical metric demonstrates inferior performance compared to the other four regression coefficient-related distance metrics. Note that the identical metric does not distinguish important dimensions from unimportant ones. Given the same value of $K$, this metric does not approximate the fully treated or controlled scenarios as effectively as other distance metrics that assign different weights to dimensions. Among the other four regression metrics, the differences are generally minimal, particularly for `Square-full'. For `Triad', however, there are subtle differences: the two Lasso-related metrics exhibit less bias compared to the two other regularization-free regression-related metrics. For $q$-frac, given that it only includes two dimensions and both are significant, all distance metrics tend to identify treated units with the highest proportion of treated neighbors, or untreated units with the highest proportion of untreated neighbors, resulting in exactly the same outcomes across all metrics.

\begin{figure}
    \centering
    \includegraphics[width=.8\linewidth]{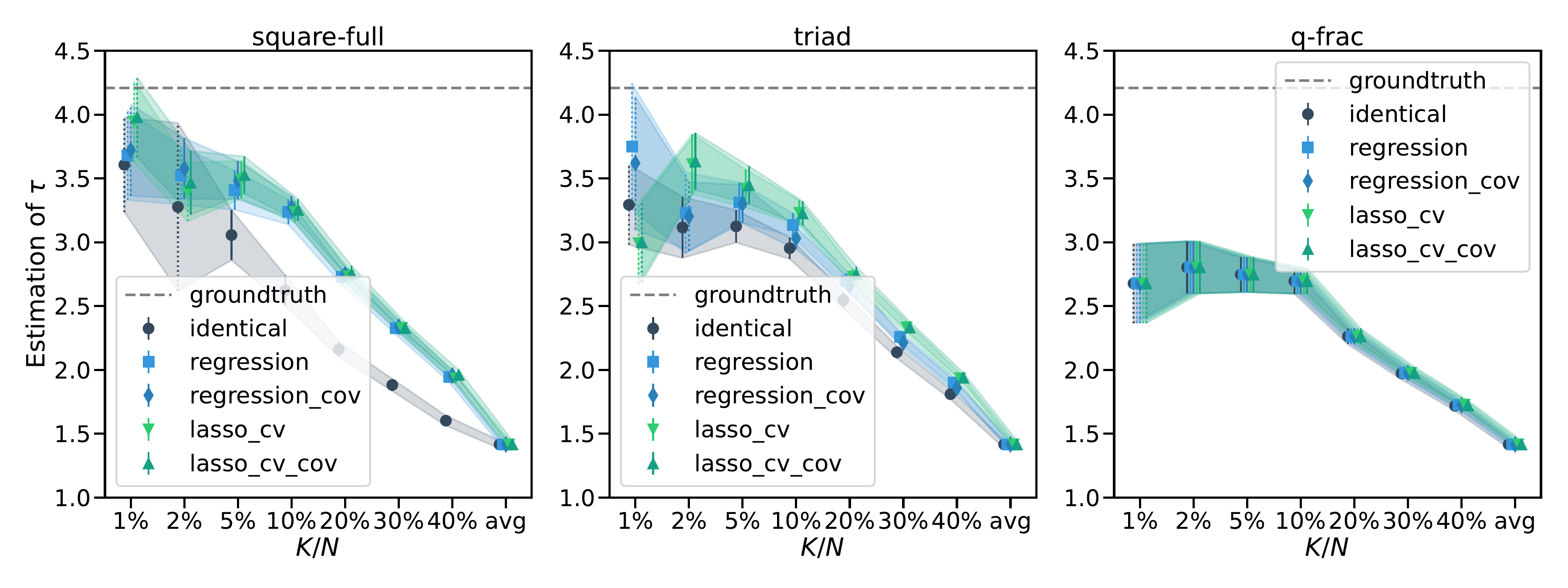}
    \caption{\raggedright\scriptsize Results for the WS synthetic experiment illustrating the base potential outcomes model across different sets of causal network motifs and distance metrics (H\'ajek estimators). Each panel depicts estimators for the global average treatment effects ($\tau$) for different sets of causal network motifs, including the benchmark ($q$-frac). Various curves represent different distance metrics. Other visual elements follow conventions in Fig.~\ref{fig:main}.
    }
    \label{fig:feature}
\end{figure}

\subsection{Augmented Model}

We propose a family of potential outcomes that extends the base model  (Eq.~\eqref{eq:base}) as follows:
\begin{equation*}
y_i(\vZ) = \frac{3}{2} (1 - \kappa + 2\kappa X_i) (1 + Z_i + \sum_{j \in \mathcal{N}_i} \lambda w_{ij} + (1-\lambda) a_{ij} Z_j) + \varepsilon_i.
\end{equation*}
Here $w_{ij} = \frac{ (1-\kappa + 2\kappa X_j)\text{cf}_{ij}}{\sum_{\ell \in \mathcal{N}_i} (1-\kappa + 2\kappa X_\ell) \text{cf}_{i\ell}} $, and 
$a_{ij} = \frac{ (1-\kappa + 2\kappa X_j)}{\sum_{\ell \in \mathcal{N}_i} (1-\kappa + 2\kappa X_\ell)} $.
When $\kappa = \frac{1}{3}$ and $\lambda=1$, this equation aligns with Eq.\eqref{eq:base}. The weight $w_{ij}$ accounts for the network structure between nodes $i$ and $j$, emphasizing edges with more common friends, whereas $a_{ij}$ treats all neighbors of $i$ equally. We anticipate that when $\lambda$ is larger, incorporating a more comprehensive set of causal network motifs, such as `Square-full', would better capture the underlying interference patterns due to its inclusion of a broader variety of network motifs, compared to a simpler set of causal network motifs like $q$-frac.
When $\kappa$ is larger, covariate-related dimensions become more important. Therefore, we expect causal network motif sets that include covariate-related dimensions (e.g., `Square-full') to outperform those that exclude them (e.g., `Square-nocov'). We have set this family of models to have the same expected global average treatment effect regardless of the values of the parameters, allowing us to directly compare their bias or variance.

To illustrate the effect of $\lambda$, we first set $\kappa=\frac{1}{3}$, consistent with the baseline model, and then vary $\lambda$. As shown in Figure~\ref{fig:lambda}, for both the `Square-full' and `Triad' models, the bias is smaller when $\lambda$ is larger, supporting our hypothesis that including complex causal network motifs is advantageous when true interference patterns involve complex network structures. By contrast, when $\lambda=0$ which suggests that network structure beyond the simple fraction of treated neighbors does not play a role, the $q$ fractional neighborhood exposure sufficiently characterizes interference. 

Moreover, we illustrate the effect of $\kappa$ in Figure~\ref{fig:kappa}. We observe that adding covariate-specific causal network motifs significantly reduces bias when $\kappa$ is large, aligning with our expectations. By contrast, when $\kappa = 0$, which indicates covariates do not affect potential outcomes, as expected we do not observe a difference between including covariate-specific causal network motifs or not.

In addition, we further explore the benefits of incorporating additional covariates and performing partialling-out in \textit{Appendix~\ref{sec:more}}. This analysis demonstrates that accounting for covariates can further reduce estimation variance.

\begin{figure}
    \centering
    \includegraphics[width=0.8\linewidth]{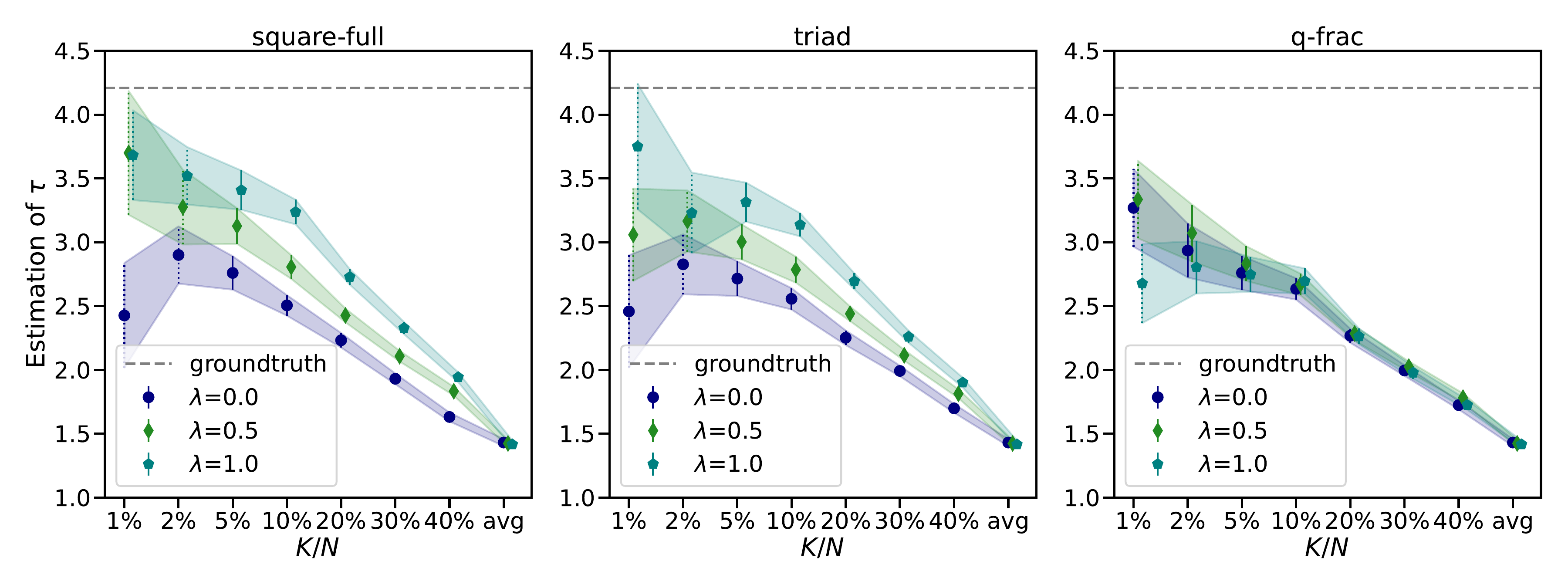}
    \caption{\raggedright \scriptsize Results for the WS synthetic experiment illustrating parameterized potential outcomes under different values for $\lambda$. Each panel depicts estimators for the global average treatment effects ($\tau$) for different sets of causal network motifs, including the benchmark ($q$-frac). Various curves within the panels represent different choices of $\lambda$. Other visual elements follow conventions in Fig.~\ref{fig:main}. }
    \label{fig:lambda}
\end{figure}

\subsection{Linear Dynamic Models and Degree of Interference }

Moreover, we adopted the linear model previously used in multiple studies, such as those \cite{eckles2016design, chin2019regression}. We consider time steps indexed by $T=1,2,3$:
\begin{equation}
    Y^*_{i, T} = \alpha + \beta Z + \gamma \sum_{j\in\mathcal{N}} w_{ij}  Y_{i, T-1} + \epsilon_i \quad\text{ and }\quad Y_{i, T} = \mathbbm{1} [Y^*_{i, T} > 2.5].\label{eq:linear}
\end{equation}
where $\epsilon_i\sim\text{Unif}([0, 1])$ and $Y$ is initialized as $Y_{i,0}=Z_i$.
Here, $T$ indicates the extent of interference. Specifically, $T=1$ signifies that only first-degree neighbors are affected. Values of $T=2$ or $T=3$ expand this interference to include second-degree and third-degree neighbors, respectively. This potential outcomes model simulates the real-world social contagion process~\citep{aral2012identifying}: Network interference from the second hop or higher is reflected when second-degree or higher neighbors initially affect the outcomes of first-degree neighbors, which in turn indirectly impacts the ego's outcomes. As an illustration, these parameters are set as follows: $\alpha = 0.5$, $\beta = 1$, $\gamma = 3$.

\begin{figure}
    \centering
    \includegraphics[width=0.8\linewidth]{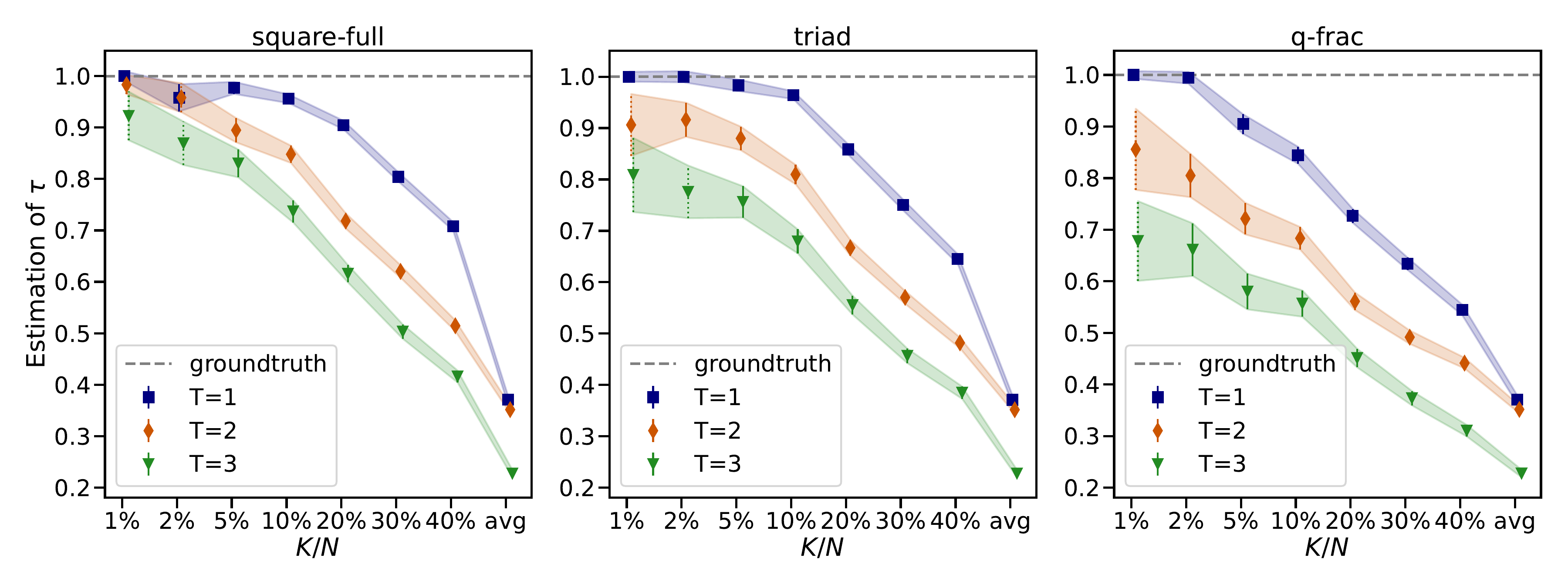}
    \caption{\raggedright \scriptsize Results for the WS synthetic experiment for illustrating linear dynamic potential outcomes under different degrees of network interference (H\'ajek estimators). Each panel depicts estimators for the global average treatment effects ($\tau$) for choices of different sets of causal network motifs. Different curves represent different choices of the degree of interference ($T$). Other visual elements follow conventions in Fig.~\ref{fig:main}.}
    \label{fig:T}
\end{figure}

We present the estimations for different \(T\) (degrees of interference) in Figure~\ref{fig:T}. We observe that the bias is larger when \(T=2\) or \(T=3\) compared to \(T=1\), primarily because all motifs considered are restricted to \(1\)-hop ego networks, thus they do not fully capture the treatments and network structure beyond 1-hop neighborhood networks. However, we also show that a smaller estimation bias is present when a small $K$ is selected regardless of $T$, suggesting that the use of a comprehensive set of causal network motifs (such as `Square-full') combined with a small $K$ can substantially reduce bias when estimating global average treatment effects. 
Furthermore, incorporating a more comprehensive set of causal network motifs does provide less biased estimation than simple approaches ($q$-frac).
In practice, although it is advisable to perform a statistical test for $n$-hop network interference such as \cite{athey2018exact}, using motifs as we have and appropriately selecting $K$ can still significantly reduce bias even if the 1-hop network interference assumption is not satisfied.

\subsection{Violation of Monotonic Interference Assumption}
We further examine scenarios where the monotonic interference assumption is violated. We revised the base model (Eq.~\eqref{eq:base}) to the potential outcomes below. 
\begin{equation*}
y_i(\vZ) = 1 + Z_i + \sum_{j \in \mathcal{N}_i}( 2 w_{ij}^{+} - w_{ij}^{-} ) Z_j + \varepsilon_i.
\end{equation*}
Here $w_{ij}^{+} = \frac{ \mathbbm{1}[X_i=X_j] (1 + X_j)\text{cf}_{ij}}{\sum_{\ell \in \mathcal{N}_i} (1+X_\ell) \text{cf}_{i\ell}} $, and 
$w_{ij}^{-} = \frac{ \mathbbm{1}[X_i \neq X_j] (1+ X_j)\text{cf}_{ij}}{\sum_{\ell \in \mathcal{N}_i} (1+X_\ell) \text{cf}_{i\ell}} $. Here, the monotonic interference assumption is not satisfied due to individuals with similar attributes ($X_i$) exerting positive interference, while those with differing attributes exert negative interference.

As depicted in Figure~\ref{fig:mono}, despite not satisfying monotonic interference, a decreasing trend is observed—lower values of $K$ yield less biased estimators, while higher values increase bias. Given that a smaller $K$ more closely approximates the fully treated or controlled scenario, opting for an estimator from a smaller $K$ provides a less biased estimation.
As a practical suggestion, unless the process of specifying the distance metric is not carefully executed, a smaller $K$ should still be preferred, provided the positivity requirement is satisfied and the variance is within reasonable bounds. This approach still well approximates the fully treated and controlled scenarios, thereby providing a more precise estimation of global average treatment effects.

\subsection{Slashdot Social Network}  We next employ the publicly available network data from Slashdot, a technology-related news website known for its specific user community \citep{leskovec2009community}. 
This network has 82,168 nodes and 582,533 edges, with a long-tailed degree distribution.  
We generate the same potential outcomes as the Watts-Strogatz network and present the estimation results in \textit{Appendix~\ref{sec:appendix:Slashdot}}. These results show that the network structure should not affect our conclusion.

\section{Real-World A/B Test}

\noindent\textbf{Instagram Tutorial Test.} Finally, to showcase the effectiveness of our approach in a real-world setting, we applied it to a randomized experiment involving 1-2 million users on the Instagram platform. This experiment aimed to enhance the user experience for a new feature called Instagram Avatars.
Instagram Avatars was introduced in February 2022 as a way for users to better express themselves in the digital world.\footnote{\url{https://www.instagram.com/p/CZe7TfGgyFU/}} 
The Avatar product allows users to create a 3D digital version of themselves and use it in stories and direct messages,  referred to as Instagram Direct. To help users better understand how to use the new Avatars product, Instagram ran an A/B test in August 2022 that experimented with a new interactive experience. The experience raises awareness about the Avatars product by testing a flow that displays all possible Avatars and shows users how to customize them. After users are shown a grid of different Avatars, it immediately prompts them to share with others. Users in the treatment group were exposed to this tutorial, which guides users on Avatar creation and facilitates sharing while users in the control group did not get the tutorial experience.

We analyzed active Instagram users between August 29 and September 27, 2022, resulting in a balanced sample of 1-2 million users. Note that all data were de-identified and analyzed in aggregate; this part of the analysis was performed internally at Meta. Users in the experiment were randomly split into treatment and control conditions under Bernoulli randomization.
We examined the impact of the Avatar tutorial on the likelihood of users using Avatars in Instagram Direct, which was coded as a binary outcome variable. When analyzing the data on September 27, 2022, between users who had been previously exposed to the tutorial and those who had not, we estimated a treatment effect of \(\hat{\tau} = 0.10\% \pm 0.03\%\). This indicates that the tutorial was effective in increasing the usage of Avatars.

However, interference may exist in this experiment which could mislead the estimation of the global average treatment effect. For instance, even if a user is in the control condition, their awareness and likelihood to use the Avatar product may increase if they have more neighbors in the treatment condition. Consequently, naively comparing treatment and control conditions may underestimate the true global average treatment effect or yield incorrect insights into the experiment's impact.
To address potential interference, we propose using causal network motif features based on the mutual follow graph among users in the A/B test on Instagram. Specifically, we employ the fraction of treated neighbors (denoted as \(2-1\)), fully treated closed triads (\(3c-2\)), and fully non-treated closed triads (\(3c-2\)) as causal network motifs to illustrate this approach.

\begin{figure}
    \centering
    \includegraphics[width=0.8\linewidth]{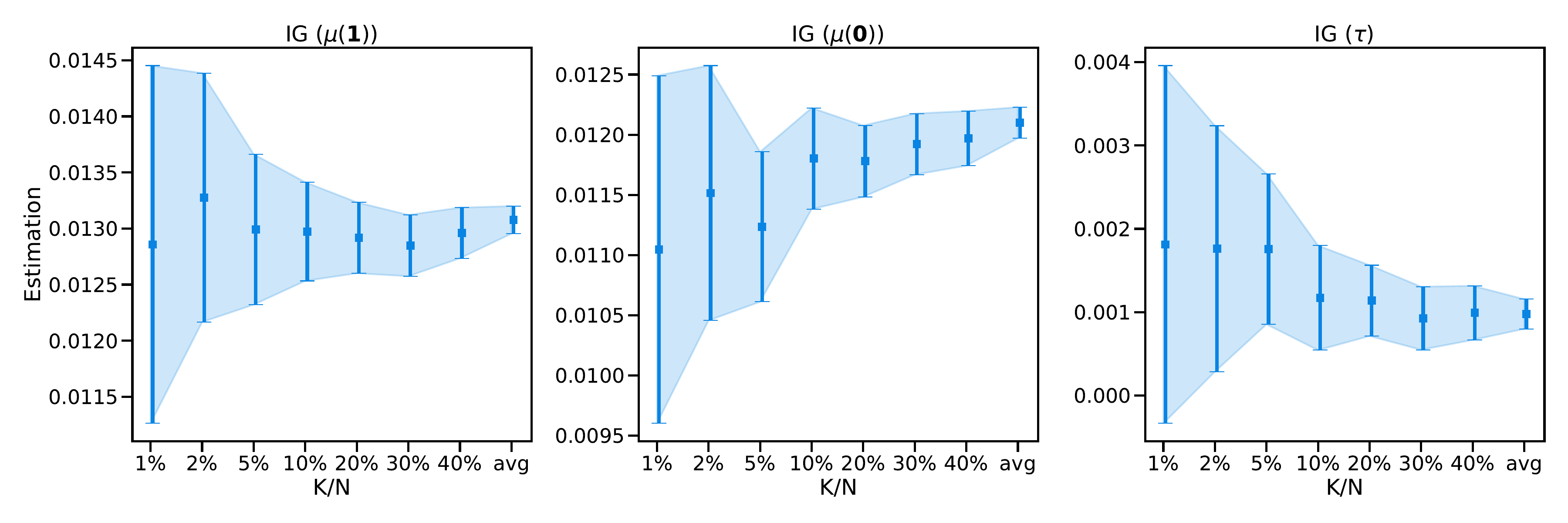}
    \caption{\raggedright\scriptsize Estimation Utilizing Nearest Neighbor Approach for Instagram Experiment under regression distance metrics: The figure showcases H\'ajek estimators employing Bernoulli randomization. Other visual elements follow conventions in Fig.~\ref{fig:main}.}
    \label{fig:IG_reg}
\end{figure}

\noindent\textbf{Estimation Result.}
We illustrate the nearest neighbor results in  Figure~\ref{fig:IG_reg}.\footnote{Here we employed a bootstrapping method to construct confidence intervals (500 replications). Briefly, for each replication \(b\), we first used sampling with replacement to regenerate the population \(\mathcal{N}^b\). We then performed estimations on this regenerated population. We reported the average and variance for these estimates. Our empirical observations indicate that this bootstrapping approach yields wider and thus more conservative confidence intervals.} Using the regression coefficients metric as an example, we observe an increasing pattern for \(\hat{\mu}(\mathbf{0})\) in \(K\) in the middle panel, ignoring the first two estimations with very large variance.
By contrast, a clear monotonic trend is not observed for the treatment group, indicating that interference may not be strong for users in this group (left panel). Based on the results when \(K/N = 5\%\) from the last panel of Figure~\ref{fig:IG_reg}, we learn that the estimation of \(\tau\) could be biased by more than 75\% if we simply contrast the averages of the treatment versus control groups (under the monotonic interference assumption). This further supports the capability of our approach to reduce bias in the presence of interference.

\section{Concluding Remarks}\label{sec:conclusions}

In summary, our study presents a two-part approach to characterizing network interference in A/B testing within networks. We build on the exposure mapping framework proposed by \cite{aronow2017estimating}, which involves specifying the exposure mapping function $f$. Our methods further decompose $f$ into two steps, represented as $f = g \cdot h$. The $g$-step models the treatments within a unit’s network neighborhood as an $M$-dimensional vector, where each dimension corresponds to a ``causal network motif.'' These motifs describe the treatment configurations of specific network structures surrounding an individual (or ego). The selection of network motifs and the determination of dimensionality $M$ draw parallels with feature engineering in standard machine learning, albeit with the goal of causal inference. The $h$-step involves mapping this $M$-dimensional vector to an exposure condition using clustering techniques—such as distance metric-based algorithms—that partition the space of causal network motif representations $[0,1]^M$ into disjoint subspaces, each corresponding to an exposure condition. We specifically explore a $K$NN-based method focused on estimating the global average treatment effect. This method targets two specific exposure conditions, representing fully treated and fully controlled scenarios. We discuss how the choice of distance metric $d$ and the number of neighbors $K$ influences the precision of causal estimates, supported by both theoretical guidelines and empirical validations through synthetic and real-world experiments. 

Our study distinguishes itself amidst the ongoing literature on analysis-based network interference methods. Primarily, our use of causal network motifs offers a practical means of defining features within the frameworks proposed by \cite{chin2019regression,awan2020almost,cortez2022exploiting,qu2021efficient,belloni2022neighborhood}, enabling these models to account for complex network structures. Furthermore, in contrast to works such as \cite{sussman2017elements,chin2019regression,cortez2022exploiting,yu2022estimating} that rely on linearity or additivity assumptions or  analytical approaches, we employ machine learning techniques, providing an alternative perspective to algorithmically characterize network interference. 
Moreover, while \cite{qu2021efficient} proposed an approach for analyzing heterogeneity in interference, a combination of their approach on accounting for neighbors' covariates with our focus on network structure potentially further improves estimation accuracy. 

Our methodology offers three promising applications for practitioners. First, integrating our approach into experimentation systems allows for a retrospective review of past A/B tests, providing insights into network interference patterns and facilitating more accurate estimation of global average treatment effects. Second, the insights (i.e., exposure conditions defined by the algorithm) obtained from our method can aid in the improvements of both design-based (e.g., redefining network structures for graph cluster randomization) and analysis-based strategies (e.g., defining more appropriate exposure mapping) for future experiments. Finally, our approach can assist in identifying the exposure conditions that yield the most beneficial outcomes, thereby enabling optimal distribution of interventions, in areas such as targeting for product promotions.

Our work suggests several paths for further exploration. First, the adaptation of our method to observational studies can be an interesting but challenging avenue particularly when randomized experiments are not viable. Existing studies have explored addressing the effect of network interference in observational settings~\citep{awan2020almost,belloni2022neighborhood}. For example, \cite{leung2022unconfoundedness} proposed a Graph Neural Network (GNN) framework. We suggest combining our approach of using causal network motifs and techniques such as double machine learning~\citep{chernozhukov2017double} to tackle network interference in observational settings.\footnote{In practice, it is not uncommon that some users are excluded from A/B testing for reasons like user agreements. One suggestion to address this issue is to frame it as an observational causal inference problem, utilizing methods like double machine learning to predict outcomes in the non-tested sample. In this context, causal network motifs can be employed as high-dimensional treatment variables, and network motifs can also be incorporated into covariates. } Moreover, our approach may be suitable for situations where only parts of a network are observable or even in non-network contexts. This flexibility may also prove useful when access to user data or specific network attributes is restricted. Finally, our approach could be integrated with the modern influence maximization problem in social networks, a field that currently emphasizes the modeling of the diffusion process, rather than placing greater focus on causality.

\begingroup
\setlength{\bibsep}{-2.pt plus -5ex}
\bibliographystyle{informs2014}
\bibliography{main}
\endgroup

\newpage

\setcounter{section}{0} 
\setcounter{figure}{0} 
\setcounter{table}{0}    

\renewcommand\thesection{\Alph{section}}    
\renewcommand\thefigure{A.\arabic{figure}}    
\renewcommand\thetable{A.\arabic{table}}

{\centering{\Large{\textbf{Appendix (Online Supplement)}}}}

\section{Supplemental Figures and Tables}
\begin{figure}[h!]
    \centering
    \includegraphics[width=0.68\linewidth]{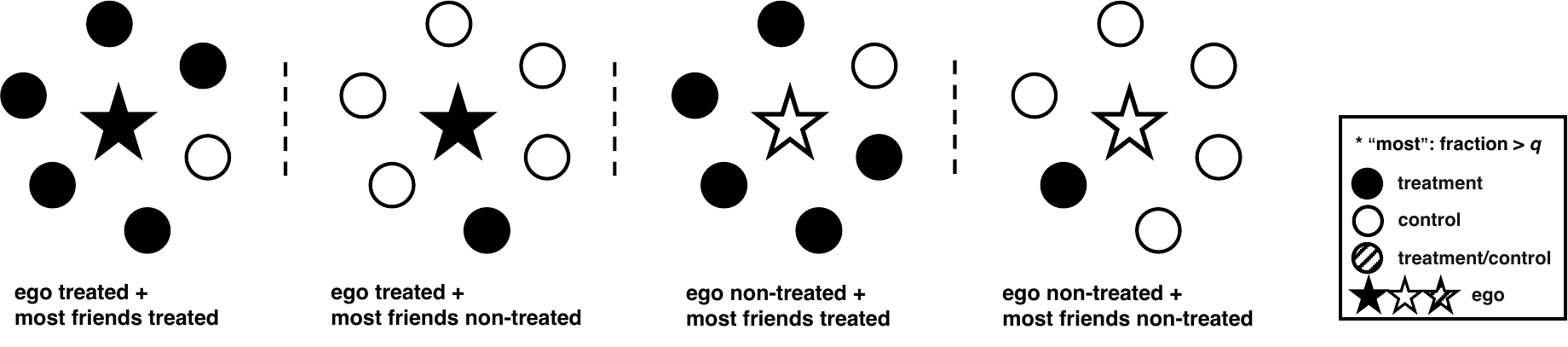}
    \caption{\raggedright\scriptsize Illustration of one possible manual specification of exposure mapping, i.e. fractional $q$-neighborhood exposure~\citep{ugander2013graph,eckles2016design}. There are four exposure conditions, 
    depending on the treatment condition of an ego node (the central nodes) and whether more than a fraction of $q$ of their network neighbors (the surrounding nodes) are treated. Red indicates treatment and yellow indicates control. 
    This specification implies that the neighbors' treatments affect the ego's outcome. }
    \label{fig:q}
\end{figure}
\begin{figure}[h!]
    \centering
    \includegraphics[width=0.48\linewidth]{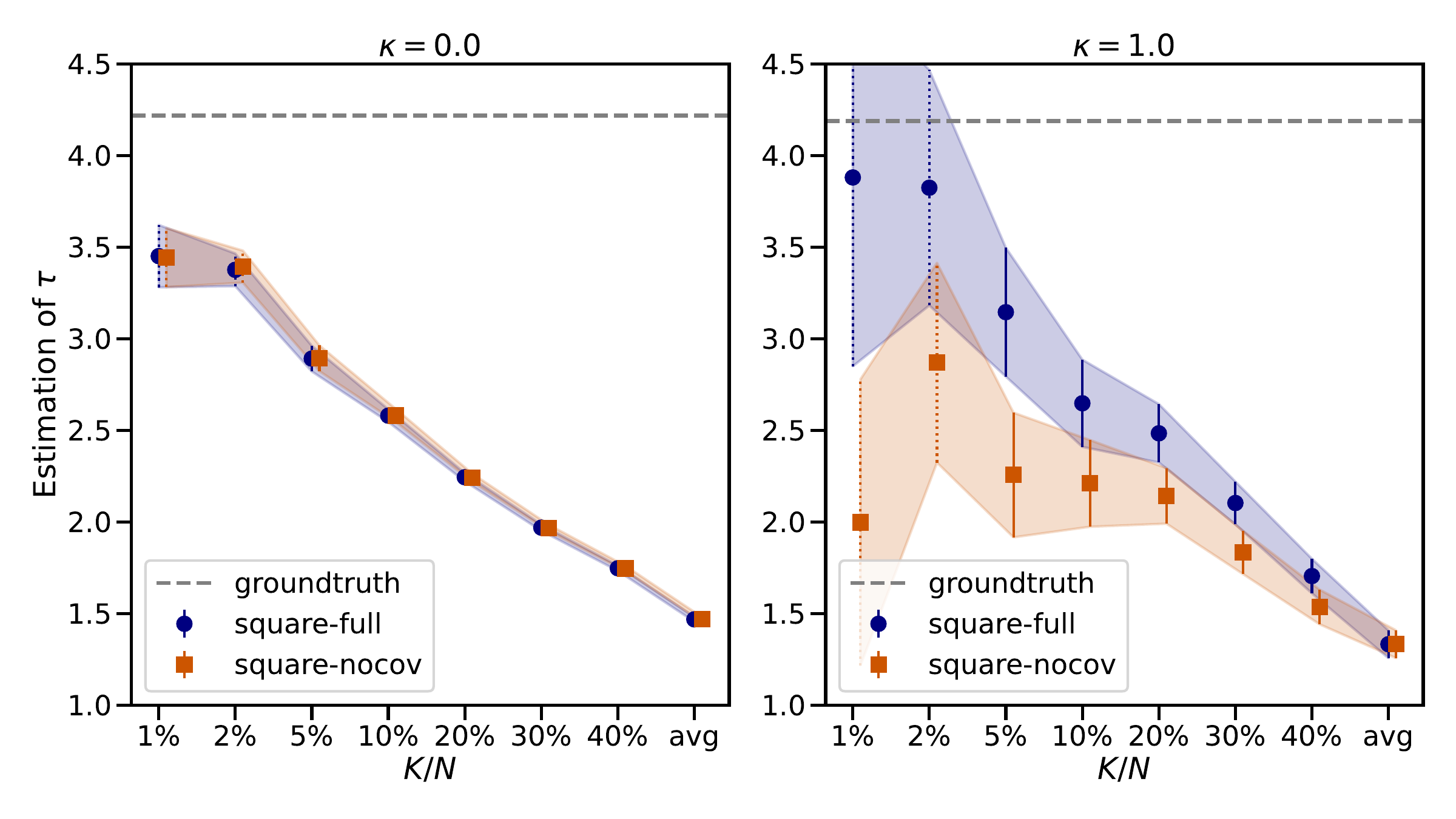}
    \caption{\raggedright\scriptsize  Results for the Watts-Strogatz synthetic experiment for the Watts-Strogatz synthetic experiment illustrating parameterized potential outcomes under different values for $\kappa$ (H\'ajek estimators). Each panel depicts estimators for the global average treatment effects ($\tau$) for choices of $\kappa$. The $x$-label $K/N$ represents the fraction of the number of nodes used in the nearest neighbor exposure condition relative to the entire population. Error bars depict standard errors, wherein the dotted error bars indicate estimates that do not meet the positivity requirement. The dashed gray lines represent the ground truth. The label ``avg" denotes the approach of calculating average outcomes for treatment and/or control groups.}
    \label{fig:kappa}
\end{figure}
\begin{figure}[h!]
    \centering
    \includegraphics[width=0.3\linewidth]{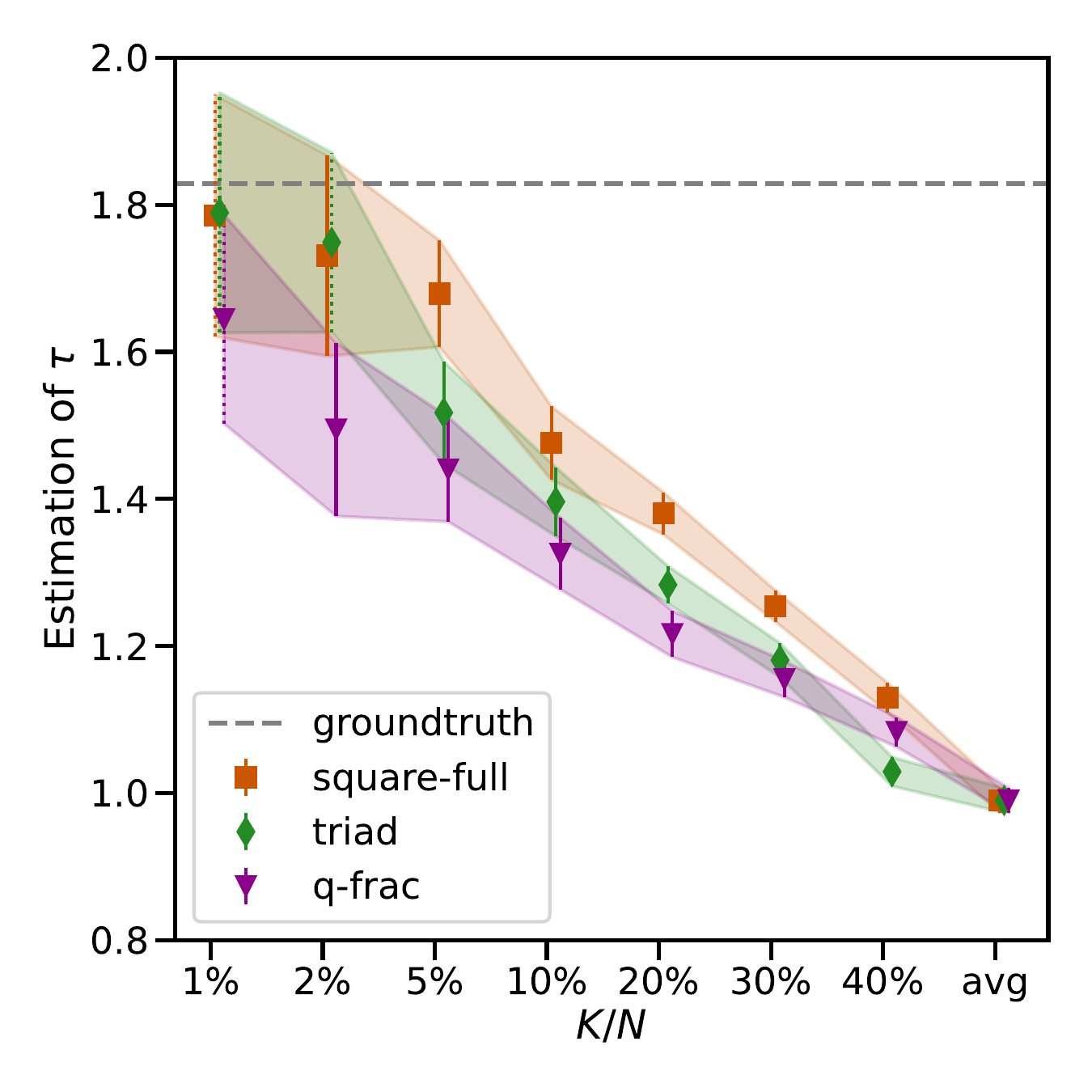}
    \caption{\raggedright \scriptsize Results for the Watts-Strogatz synthetic experiment for illustrating violation of the monotonic interference models (H\'ajek estimators). Different curves represent different sets of causal network motifs. The $x$-label $K/N$ represents the fraction of the number of nodes used in the nearest neighbor exposure condition relative to the entire population. Error bars depict standard errors, wherein the dotted error bars indicate estimates that do not meet the positivity requirement. The dashed gray lines represent the ground truth. The label ``avg" denotes the approach of calculating average outcomes for treatment and/or control groups.}
    \label{fig:mono}
\end{figure}

\newpage
\section{Proofs}
\subsection{Proof for Lemma~\ref{eq:lemma:est}}\label{sec:lemma:est}
\prf{
\begin{equation*}
\begin{split}
\Exp[\hat{\mu}_{\text{HT}}^f(\delta)] &= 
\frac{1}{N} \sum_{i\in\mathcal{N}} \Exp_{\vZ}\left[ \frac{Y_i \mathbbm{1}[ f(\vZ, \theta_i) = \delta ]}{\mathbb{P}[f(\vZ, \theta_i) = \delta]} \right] 
\\ & = \frac{1}{N}  \sum_{i\in\mathcal{N}} \sum_{\vz \in \{0,1\}^N} \frac{y_i(\vz)  \mathbbm{1}[ f(\vz, \theta_i) = \delta ] \mathbb{P}[\vZ=\vz]}{  \mathbb{P}[f(\vz, \theta_i) = \delta]} 
\\ & = \frac{1}{N} \sum_{i\in\mathcal{N}} \frac{y_i(\vz)  \mathbbm{P}[ f(\vz, \theta_i) = \delta \cap \vZ=\vz]}{  \mathbb{P}[f(\vz, \theta_i) = \delta]} 
\\ &= \frac{1}{N} \Exp_{\vZ}[ y_i(\vZ) | f(\vz, \theta_i) = \delta ] = \frac{1}{N} {\mu}^f_i(\delta) = {\mu}^f(\delta).
\end{split}
\end{equation*}
}

\subsection{Proof for Proposition~\ref{prop:misspecify}}\label{proof:misspecify}
\prf{
\begin{enumerate}
\item \begin{align*}
\tau^f = \Exp_{\vZ} [\hat\tau_{\text{HT}}^f] 
& = \frac{1}{N} \Exp_{\vZ} \left[ \frac{\sum_{i\in\mathcal{N}} Y_i \mathbbm{1}[ f(\vZ, \theta_i) = \delta^{(1)} ]}{\mathbb{P}[f(\vZ, \theta_i) = \delta^{(1)}]} - \frac{\sum_{i\in\mathcal{N}} Y_i \mathbbm{1}[ f(\vZ, \theta_i) = \delta^{(0)} ]}{\mathbb{P}[f(\vZ, \theta_i) = \delta^{(0)}]} \right] \\
& = \frac{1}{N} \Exp_{\vZ} \left[ \frac{\sum_{i\in\mathcal{N}} \mu_i^f(  \delta^{(1)}  ) \mathbbm{1}[ f(\vZ, \theta_i) = \delta^{(1)} ]}{\mathbb{P}[f(\vZ, \theta_i) = \delta^{(1)}]} - \frac{\sum_{i\in\mathcal{N}} \mu_i^f(  \delta^{(0)}  ) \mathbbm{1}[ f(\vZ, \theta_i) = \delta^{(0)} ]}{\mathbb{P}[f(\vZ, \theta_i) = \delta^{(0)}]}  \right] 
\\ & =  \Exp_{\vZ} \left[ \frac{\sum_{i\in\mathcal{N}} \mu_i^f(\delta^{(1)}) w_i^f(\delta^{(1)})}{N} - \frac{\sum_{i\in\mathcal{N}} \mu_i^f(\delta^{(0)}) w_i^f(\delta^{(0)})}{N}\right]
\\ & =  \Exp_{\vZ} \left[ \frac{\sum_{i\in\mathcal{N}} y_i(\bm{1}) w_i^f(\delta^{(1)})}{N} - \frac{\sum_{i\in\mathcal{N}} y_i(\bm{0}) w_i^f(\delta^{(0)})}{N}\right]
\\ & =   \frac{\sum_{i\in\mathcal{N}} y_i(\bm{1}) \Exp_{\vZ} \left[ w_i^f(\delta^{(1)}) \right]}{N} - \frac{\sum_{i\in\mathcal{N}} y_i(\bm{0}) \Exp_{\vZ} \left[ w_i^f(\delta^{(0)})\right]}{N}
\\ & = \frac{1}{N} \sum_i \left( y_i(\bm{1}) - y_i(\bm{0}) \right)  =  \tau 
\end{align*}
The third and fifth equalities are derived from the assumption that the exposure mapping is correctly specified.
\item 
\begin{align*}
\tau^f =\Exp_{\vZ} [\hat\tau_{\text{HT}}^f] 
& = \Exp_{\vZ} \left[ \frac{\sum_i Y_i w_i^f(\delta^{(1)})}{N} - \frac{\sum_i Y_i w_i^f(\delta^{(0)})}{N} \right] 
\\ & =  \Exp_{\vZ} \left[ \frac{\sum_i y_i(\bm{Z}) w_i^f(\delta^{(1)})}{N} - \frac{\sum_i y_i(\bm{Z}) w_i^f(\delta^{(0)})}{N} \right]
\\   & \leq  \Exp_{\vZ} \left[ \frac{\sum_i y_i(\bm{1}) w_i^f(\delta^{(1)})}{N} - \frac{\sum_i y_i(\bm{0}) w_i^f(\delta^{(0)})}{N} \right] 
\\ & = \frac{1}{N} \sum_i \left( y_i(\bm{1}) - y_i(\bm{0}) \right)  =  \tau 
\end{align*}
The inequality is from the non-negative interference assumption. The fourth equality is from the fact that $\Exp_{\vZ} \left[\frac{\mathbbm{1}[ f(\vZ, \theta_i) = \delta  ]}{\mathbbm{P}[ f(\vZ, \theta_i) = \delta]} \right] = 1$ for $\delta=\delta^{(1)}$ and $\delta^{(0)}$.
\item This is symmetric to \textit{proof}~2 except that  the sign should be ``$\geq$'' and non-positive interference should be assumed. 
\end{enumerate}
}

\subsection{Proof for Proposition~\ref{prop:K}}
\label{appendix:K}
\prf{
Here we only provide detailed proof for non-negative interference assumption and the proof for non-positive interference assumption is symmetric. 

Consider two possible values for $K$: $\bar{K}$ and $\ubar{K}$ ($\ubar{K} \leq \bar{K}$). Note that we should only consider $K$ that lead to exposure conditions that satisfy the positivity requirement. 
Let $\mathcal{R}_{d,\bar{K}}^{(1)}$ and $\mathcal{R}_{d,\bar{K}}^{(0)}$
denote the exposure conditions corresponding to the fully treated or non-treated scenarios defined by the $\bar{K}$ nearest neighbors. 
Let $\mathcal{R}_{d,\ubar{K}}^{(1)}$ and $\mathcal{R}_{d,\ubar{K}}^{(0)}$
denote the exposure conditions corresponding to the fully treated or non-treated scenarios defined by the $\ubar{K}$ nearest neighbors.  
\begin{equation*}
\begin{split}
    \Exp_{\vZ}[\hat{\tau}_{\text{HT}}^{d,\bar{K}}] - \Exp_{\vZ}[\hat{\tau}_{\text{HT}}^{d,\ubar{K}}]  = \frac{1}{N} \sum_{i\in\mathcal{N}} \left( \Exp_{\vZ} [y_i(\vZ) | g(\vZ, \theta_i) \in \mathcal{R}_{d,\bar{K}}^{(1)} ] - \Exp_{\vZ} [y_i(\vZ) | g(\vZ, \theta_i) \in \mathcal{R}_{d,\ubar{K}}^{(1)} ] \right) \\ - \frac{1}{N} \sum_{i\in\mathcal{N}} \left( \Exp_{\vZ} [y_i(\vZ) | g(\vZ, \theta_i) \in \mathcal{R}_{d,\bar{K}}^{(0)} ] - \Exp_{\vZ} [y_i(\vZ) | g(\vZ, \theta_i) \in \mathcal{R}_{d,\ubar{K}}^{(0)} ] \right)
\end{split}
\end{equation*}
We know $\mathcal{R}_{d,\ubar{K}}^{(1)} \subseteq \mathcal{R}_{d,\bar{K}}^{(1)}$ according to the nature of nearest neighbors algorithm, i.e. we would include more units but do not exclude as we increase $K$. 
Therefore we can separate $\mathcal{R}_{d,\bar{K}}^{(1)}$ into $\mathcal{R}_{d,\ubar{K}}^{(1)}$ and $\mathcal{R}_{d,\bar{K}}^{(1)} \setminus \mathcal{R}_{d,\ubar{K}}^{(1)}$ and then we derive:
\begin{align}
    & \frac{1}{N} \sum_{i\in\mathcal{N}} \left( \Exp_{\vZ} [y_i(\vZ) | g(\vZ, \theta_i) \in \mathcal{R}_{d,\bar{K}}^{(1)} ] - \Exp_{\vZ} [y_i(\vZ) | g(\vZ, \theta_i) \in \mathcal{R}_{d,\ubar{K}}^{(1)} ] \right) \\ = & \frac{1}{N} \sum_{i\in\mathcal{N}} \left( p \Exp_{\vZ} [y_i(\vZ) | g(\vZ, \theta_i) \in \mathcal{R}_{d,\ubar{K}}^{(1)} ]  + (1-p) \Exp_{\vZ} [y_i(\vZ) | g(\vZ, \theta_i) \in \mathcal{R}_{d,\bar{K}}^{(1)} \setminus \mathcal{R}_{d,\ubar{K}}^{(1)} ]  
     - \Exp_{\vZ} [y_i(\vZ) | g(\vZ, \theta_i) \in \mathcal{R}_{d,\ubar{K}}^{(1)} ] \right) \\
      = & \frac{1}{N} \sum_{i\in\mathcal{N}}  (1-p) \left( \Exp_{\vZ} [y_i(\vZ) | g(\vZ, \theta_i) \in \mathcal{R}_{d,\bar{K}}^{(1)} \setminus \mathcal{R}_{d,\ubar{K}}^{(1)} ]  
     -  \Exp_{\vZ} [y_i(\vZ) | g(\vZ, \theta_i) \in \mathcal{R}_{d,\ubar{K}}^{(1)} ]  \right) 
\label{eq:label:ineq}
\end{align}
\noindent where $\Pro[g(\vZ, \theta_i) \in \mathcal{R}_{d,\ubar{K}}^{(1)} | g(\vZ, \theta_i) \in \mathcal{R}_{d,\bar{K}}^{(1)}] = p$ and $\Pro[ g(\vZ, \theta_i) \in \mathcal{R}_{d,\bar{K}}^{(1)} \setminus \mathcal{R}_{d,\ubar{K}}^{(1)} | g(\vZ, \theta_i) \in \mathcal{R}_{d,\bar{K}}^{(1)}] = 1-p$. The first equality follows the law of total expectation. The second quality is derived from merging the first and third terms. 

Again by the nature of nearest neighbor algorithm and given the distance metric $d$, for all $\vr \in \mathcal{R}_{d,\bar{K}}^{(1)} \setminus \mathcal{R}_{d,\ubar{K}}^{(1)} $ and $\vr' \in \mathcal{R}_{d,\ubar{K}}^{(1)} $, $d(\vr^{(1)}, \vr) \geq  d(\vr^{(1)}, \vr')$. This is because the $\ubar{K}$ nearest neighbors of $\vr^{(1)}$ should be a subset of the $\bar{K}$ nearest neighbors of $\vr^{(1)}$.
As we assume that $d$ is properly specified, we can convert the inequality of distance metric $d$ to the potential outcome $y_i$, i.e.,
\begin{equation}
\Exp_{\vZ} [y_i(\vZ) | g(\vZ, \theta_i) \in\mathcal{R}_{d,\bar{K}}^{(1)} \setminus \mathcal{R}_{d,\ubar{K}}^{(1)} ]  
     \leq  \Exp_{\vZ} [y_i(\vZ) | g(\vZ, \theta_i) \in \mathcal{R}_{d,\ubar{K}}^{(1)} ].
\end{equation}
    
Then by combining this with Eq.~(\ref{eq:label:ineq}), we have
    \begin{equation}
        \frac{1}{N} \sum_{i\in\mathcal{N}} \Exp_{\vZ} [y_i(\vZ) | g(\vZ, \theta_i) \in \mathcal{R}_{d,\bar{K}}^{(1)} ] \leq \frac{1}{N} \sum_{i\in\mathcal{N}} \Exp_{\vZ} [y_i(\vZ) | g(\vZ, \theta_i) \in \mathcal{R}_{d,\ubar{K}}^{(1)} ] .
    \end{equation}
    Similarly, we can derive \begin{equation}
        \frac{1}{N} \sum_{i\in\mathcal{N}} \Exp_{\vZ} [y_i(\vZ) | g(\vZ, \theta_i) \in \mathcal{R}_{d,\bar{K}}^{(0)} ] \geq \frac{1}{N} \sum_{i\in\mathcal{N}} \Exp_{\vZ} [y_i(\vZ) | g(\vZ, \theta_i) \in \mathcal{R}_{d,\ubar{K}}^{(0)} ] .
    \end{equation}
    Therefore, we can conclude $\Exp_{\vZ}[\hat{\tau}_{\text{HT}}^{d,\bar{K}}] \leq \Exp_{\vZ}[\hat{\tau}_{\text{HT}}^{d,\ubar{K}}]$.
Then $\Exp_{\vZ}[\hat{\tau}_{\text{HT}}^{d,{K}}]$  decreases in $K$. 

Moreover, we learn from Proposition~\ref{prop:misspecify} that $\Exp_{\vZ}[\hat{\tau}_{\text{HT}}^{d,\bar{K}}] \leq \tau$ and 
$\Exp_{\vZ}[\hat{\tau}_{\text{HT}}^{d,\ubar{K}}] \leq \tau$ under the non-negative interference assumption. 
Thus, $|\Exp_{\vZ}[\hat{\tau}_{\text{HT}}^{d,\bar{K}}] - \tau | \geq |\Exp_{\vZ}[\hat{\tau}_{\text{HT}}^{d,\ubar{K}}] - \tau|$, i.e. the bias of  $\Exp_{\vZ}[\hat{\tau}_{\text{HT}}^{d,{K}}]$ increases in $K$.}

\section{Additional Details for Estimating General Probability of Exposure}
\label{sec:appendix:general}
As mentioned in the main text, one challenge is to compute the probability of a unit belonging to a specific exposure condition, referred to as the ``general probability of exposure.''
This probability is used in both the Horvitz–Thompson estimator and H\'ajek estimator.
We use Monte Carlo simulation to estimate this probability.  Specifically, we randomly assign treatments ($\bm{Z}$) for $B$ replicates ($B=1000$ in practice), which help generate the empirical probability of the event that $\bm{R}_i$ belongs to $\mathcal{R}$, that is $\bm{R}_i^{(1)}, \bm{R}_i^{(1)}, \cdots, \bm{R}_i^{(B)}$.
That is for unit $i$, 
\begin{equation*}
\hat{\mathbbm{P}}[\bm{R}_i \in \mathcal{R}] = \frac{\sum_{b=1}^{B} \mathbbm{1}[\bm{R}_i^{(b)} \in \mathcal{R}] }{B+1}.
\end{equation*}
The ``$+1$'' in the denominator is to avoid zero-valued denominators when we use H\'ajek or Horvitz-Thompson estimators.\footnote{This equation has shown to converge to the true probability ${\mathbbm{P}}[\bm{R}_i \in \mathcal{R}]$. See details in \cite{aronow2017estimating}.}

\section{Additional Details for Variance Estimation}

\label{sec:theoretical}

Our main text outlined the equations for variance estimators for both the average potential outcome $\mu(\cdot)$ and the global average treatment effect $\tau$.  Here we continue this discussion.
For the global average treatment effect, we mentioned that we need to estimate the covariance of the average potential outcomes in the two subspaces ($\mathcal{R}^{(1)}$) and ($\mathcal{R}^{(0)}$). Based on \cite{aronow2017estimating} the estimand is:
\begin{align*}
& \text{Cov}\left(\hat{\mu}_{\text{HT}}(\mathcal{R}^{(1)}), \hat{\mu}_{\text{HT}}(\mathcal{R}^{(0)})\right) \\ = & \sum_{\substack{i\neq j \in \mathcal{N}: \\\mathbb{P}[\vR_i\in \mathcal{R}^{(1)},\vR_j \in \mathcal{R}^{(0)}]>0}}  \frac{\mu_i(\mathcal{R}^{(1)}) \mu_j(\mathcal{R}^{(0)}) \left( \mathbb{P}[\vR_i \in \mathcal{R}^{(1)}, \vR_j \in \mathcal{R}^{(0)}]  
    - \mathbb{P}[\vR_i \in \mathcal{R}^{(1)}] \mathbb{P}[\vR_j \in \mathcal{R}^{(0)}] \right)}{ \mathbb{P}[\vR_i \in \mathcal{R}^{(1)}] \mathbb{P}[\vR_j \in \mathcal{R}^{(0)}]} \\
    &  - \sum_{\substack{i\neq j \in \mathcal{N}:\\\mathbb{P}[\vR_i\in \mathcal{R}^{(1)},\vR_j \in \mathcal{R}^{(0)}]=0}} \left(\mu_i(\mathcal{R}^{(1)} ) \mu_j(\mathcal{R}^{(0)})\right).
\end{align*}
and its estimator of it is
\begin{align}
\widehat{\text{Cov}}\left(\hat{\mu}_{\text{HT}}(\mathcal{R}^{(1)}), \hat{\mu}_{\text{HT}}(\mathcal{R}^{(0)})\right) = & \sum_{\substack{i\neq j \in \mathcal{N}: \vR_i\in\mathcal{R}^{(1)},\vR_j\in\mathcal{R}^{(0)}\\\text{ and }\mathbb{P}[\vR_i\in \mathcal{R}^{(1)},\vR_j \in \mathcal{R}^{(0)}]>0}}  \frac{ Y_i Y_j \left( \mathbb{P}[\vR_i \in \mathcal{R}^{(1)}, \vR_j \in \mathcal{R}^{(0)}]  
    - \mathbb{P}[\vR_i \in \mathcal{R}^{(1)}] \mathbb{P}[\vR_j \in \mathcal{R}^{(0)}] \right)}{ \mathbb{P}[\vR_i \in \mathcal{R}^{(1)}] \mathbb{P}[\vR_j \in \mathcal{R}^{(0)}] \mathbb{P}[\vR_i\in \mathcal{R}^{(1)},\vR_j \in \mathcal{R}^{(0)}] } \label{eq:pairwise_ind}\\
    &  - \sum_{\substack{i\neq j \in \mathcal{N}: \vR_i\in\mathcal{R}^{(1)},\vR_j\in\mathcal{R}^{(0)}\\\text{ and }\mathbb{P}[\vR_i\in \mathcal{R}^{(1)},\vR_j \in \mathcal{R}^{(0)}]=0}} \frac{1}{2} \left(\frac{Y_i^2}{\mathbb{P}[\vR_i \in \mathcal{R}^{(1)}]}+\frac{Y_j^2}{\mathbb{P}[\vR_j \in \mathcal{R}^{(0)}]} \right).
\end{align}

According to \cite{aronow2017estimating}, HT estimators are conservative variance estimators, i.e.,
\begin{enumerate}
\item $\mathbb{E}_{\vZ}\left[\widehat{\text{Var}}(\hat{\mu}_{\text{HT}}\left(\mathcal{R}_k )\right)\right]   \geq {\text{Var}}\left(\hat{\mu}_{\text{HT}}(\mathcal{R}_k)\right)$.
\item $\mathbb{E}_{\vZ}\left[\widehat{\text{Var}}\left(\hat{\tau}^{d,K}_{\text{HT}} \right) \right] \geq {\text{Var}}\left(\hat{\tau}^{d,K}_{\text{HT}} \right)$.
\end{enumerate}

\cite{aronow2017estimating} discusses the consistency of the estimator, which can directly apply to our setting: 
\begin{proposition}[Consistency; Proposition 6.1 of \cite{aronow2017estimating}]
Consider a sequence of finite populations indexed by $N$, and define $\mathcal{N}_N$ accordingly. Under these conditions:
\begin{enumerate}
\item The potential outcomes and inverse propensity scores are bounded: for all $i$, for 
$\vz \in \{0,1\}^N$, $\left| y_i(\mathbf{z}) \right| \leq C_1 < \infty$ and $\left| \frac{1}{\mathbb{P}[\mathbf{R}_i\in \mathcal{R}]} \right| \leq C_2 < \infty$ where $C_1$ and $C_2$ are constants;
\item  $\sum_{i,j \in \mathcal{N}_N} \Phi(i,j) = o(N^2)$, where $\Phi(i,j) = 1$ if 
$\vR_i \indep \vR_j$ and $\Phi(i,j) = 0$ otherwise.
\end{enumerate}
Then we have $\hat{\mu}(\mathcal{R}) - \mu(\mathcal{R}) \rightarrow 0$ as $N \rightarrow \infty$.
\end{proposition}
Accordingly, we also have $\hat{\tau}_{\text{HT}} = \left( \hat{\mu}_{\text{HT}}(\mathcal{R}^{(1)}) - \hat{\mu}_{\text{HT}}(\mathcal{R}^{(0)}) \right)
-  \left( {\mu}(\mathcal{R}^{(1)}) - {\mu}(\mathcal{R}^{(0)}) \right)
\rightarrow 0$ as $N \rightarrow \infty$.

\noindent\textbf{Discussion on Pairwise Dependency.} Estimating variance and covariance requires calculating the joint probability of any two nodes being in the same or different exposure conditions. This calculation has a time complexity of \(O(N^2)\). To address this, we recommend focusing solely on those pairs that are not independent. 
Note that under the Bernoulli randomization with \(n\)-hop network interference assumption, we have
\[
\vR_i \indep \vR_j \text{ if } j \notin \mathcal{N}_i^{2n}.
\]
This implies that the randomization vectors \(\vR_i\) and \(\vR_j\) are independent if units \(i\) and \(j\) are more than \(2n\) hops apart (therefore $\mathbb{P}[\vR_i \in \mathcal{R}^{(1)}, \vR_j \in \mathcal{R}^{(0)}]  
    - \mathbb{P}[\vR_i \in \mathcal{R}^{(1)}] \mathbb{P}[\vR_j \in \mathcal{R}^{(0)}] = 0$ in Eq.~\eqref{eq:pairwise_ind}). We can thus leverage the sparsity of real-world networks to substantially reduce the number of pairs for which we need to consider the joint probability during variance estimation. For instance, with \(n = 1\), each unit may have only a small number of second-degree neighbors due to the sparsity of real-world networks, significantly improving computational efficiency during variance estimation.

In the scenario of cluster randomization, we have
\[
\vR_i \indep \vR_j \text{ if } j' \notin \mathcal{N}_{i'}^{2n} \text{ for all } i', j' \text{ s.t. $i'$ and $i$ ($j'$ and $j$) are in the same cluster}.
\]
Here, the cluster sizes are suggested to be sufficiently small (\(\bar{C} = o(N)\)). This helps reduce the dependency between units across different clusters, and thus a majority of dependency is contained within the same clusters. Again, this allows us to improve computational efficiency during variance estimation as we do not need to compute the joint probability of many independent pairs.
Real-world networks can typically be reasonably partitioned into clusters that are relatively disconnected. This property allows us to identify a sufficient number of pairs of nodes that are independent, thus improving computational efficiency.

In addition to converge, \cite{aronow2017estimating} also proposed asymptotic validity of confidence intervals. We summarize their conclusion as follows:
 
\begin{proposition}[Asymptotic validity of CIs; Proposition 6.2 of \cite{aronow2017estimating}]
Consider a sequence of populations indexed by $N$, and define $\mathcal{N}_N$ accordingly. Under these conditions:
\begin{enumerate}
\item For all $i$, and $\vz \in \{0,1\}^N$ $\left| y_i(\mathbf{z}) \right| \leq C_1 < \infty$ and $\left| \frac{1}{\mathbb{P}[\vR_i \in \mathcal{R}]} \right| \leq C_2 < \infty$.
\item There exists a finite constant $C_3$ such that for all $\mathcal{N}_N$, and for all $i \in \mathcal{N}_N$, $\sum_{j \in \mathcal{N}_N} \Psi(i,j) \leq C_3$. Here $\Psi(i,j)$ indicates a strong condition of dependency (see Condition 5 in \cite{aronow2017estimating}).
\item $N \cdot \text{Var}[\widehat{\tau_\text{HT}}] \rightarrow C_4$, as $N \rightarrow \infty$, where $C_4 > 0$.
\end{enumerate}
If confidence intervals are constructed as:
$$\widehat{\tau_{HT}} \pm z_{1-\alpha/2} \sqrt{\widehat{\text{Var}} [\widehat{\tau_{HT}} ]},$$ 
they cover the global average treatment effect $\tau$ at least $100(1-\alpha)\%$ of the time.
\label{prop:cis}
\end{proposition}
Here the condition 2 in the Proposition~\ref{prop:cis} implies that the number of dependencies for each unit \(i\) must be \(o(N)\). 

Note that so far we discuss HT estimators only. As for H\'ajek estimators, we perform Delta methods where all elements within this approach can be derived along with the HT estimators. This is consistent with many previous work including \cite{aronow2017estimating,eckles2016design}.
\section{The Tree-Based Algorithm}

\label{appendix:tree}

\subsection{Tree-Based Approach}
\label{sec:tree} 
We use decision trees to partition the space of the causal network motif representation, which is $[0,1]^M$, into multiple disjoint subspaces: $[0, 1]^M = \mathcal{R}_1 \cup \mathcal{R}_2 \cup \mathcal{R}_3  \cup \cdots \cup \mathcal{R}_{|\Delta|}$. We adapt the ``exposure mapping'' proposed by \cite{aronow2017estimating}, where units are categorized into a number of ``leaves'' in the decision tree. However, compared with conventional decision tree regression, we need to propose the following revisions. As this was part of the main contribution of the previous conference version~\citep{yuan2021causal}, we only list the core ideas:\footnote{A related work by \cite{bargagli2020heterogeneous} extends the honest causal tree approach \citep{athey2016recursive} to explore the heterogeneous spillover effects based on a node's network features. However, the purpose of the tree-based approach in our study is completely distinct: We design our tree-based method to partition the `treatment space', with each dimension of the treatment vector reflecting a varying treatment assignment. In contrast, their approach centers on the covariate space, featuring constant network properties regardless of random assignments.
}

The algorithm is implemented by recursion (see Algorithm~\ref{algo:tree}). Specifically, we have a procedure
\textsc{Split} which is used to partition a given space in $[0, 1]^{M}$. 
One can use \textsc{Split}($[0, 1]^{M}$) to start the recursion algorithm. 
When the algorithm terminates, each leaf corresponds to an exposure condition ($\mathcal{R}$) and we then calculate $\hat\mu(\mathcal{R})$, the average potential outcome for an exposure condition $\mathcal{R}$.

In addition to estimating the average potential outcome, our approach can also be used to estimate the global average treatment effects. 
Since the mapping for causal network motif representation ($g$) is representation invariant to $\bm{z}=\bm{1}$ and $\bm{0}$, we can define $\vr^{(1)} =g(\bm{1}, \cdot)$ and $ \vr^{(0)} =g(\bm{0}, \cdot)$ for all $i \in \mathcal{N}$. Then we just need to find the two leaves of the decision trees (i.e., exposure conditions) that contain $\vr^{(1)}$ and $\vr^{(0)}$ and compare the difference between
$\hat\mu(\mathcal{R}^{(1)})$ and $\hat\mu(\mathcal{R}^{(0)})$ to estimate the global average treatment effect.

Here we discuss technical details for the tree-based algorithm:

\begin{enumerate}
    \item \textit{(Positivity in empirical general probability of exposure)}. Positivity ensures that all units have a non-zero probability of being in an exposure condition \citep{rubin2005causal}.
    The tree algorithm should not split a node if such splitting would lead to any child node (corresponding to an exposure condition) where there is a unit that violates the positivity requirement. Since $\mathbb{P}[\vR_i \in \mathcal{R}]$ is often not solvable analytically, we use Monte Carlo to approximate it (denoted by $\hat{\mathbb{P}}[\vR_i \in \mathcal{R}]$).  Moreover, we adjust the positivity requirement to non-trivial probability, which allows very few units to have zero or near-zero probability; positive real numbers $C$ and $\epsilon$ introduce a small bias but allow partitioning more features. 
    \begin{equation}
        \frac{1}{N} \sum_{i \in \mathcal{N}} \mathbbm{1} [\hat{\mathbb{P}}[\vR_i \in \mathcal{R}] \leq \epsilon ]  \leq C
        \label{eq:pos}
    \end{equation}\noindent It means that the fraction of units with $\hat{\mathbb{P}}[\vR_i \in \mathcal{R}] \leq \epsilon$ is smaller than a small constant $C$.\footnote{It is empirically advised to  set $\epsilon = 0$ and $C = 0.01$.}

        \item \textit{(Splitting rule)} Here we introduce two score functions for the tree algorithm to determine when to further split given a current node (represented by $\mathcal{R}$) into its left and right children (denoted by $\mathcal{R}_{\text{left}}$ and $\mathcal{R}_{\text{right}}$ respectively): $t$ statistic and the weighted sum of squares error (WSSE). 
    
    For the $t$ statistic, we define 
    \begin{equation*}
        \text{Score}_t(\mathcal{R}_{\text{left}}, \mathcal{R}_{\text{right}}) = \frac{| \hat{\mu}(\mathcal{R}_{\text{left}}) - \hat{\mu}(\mathcal{R}_{\text{right}}) | }{ \sqrt{ \widehat{\text{Var}}  \left( \hat{\mu}(\mathcal{R}_{\text{left}}) - \hat{\mu}(\mathcal{R}_{\text{right}}) \right)} }
    \end{equation*}

Here $\hat{\mu}$ can be either H\'ajek or HT estimator. It is essential the $t$ statistic for the test for the null hypothesis that the average potential outcomes for the two resulting exposure conditions are equal. A large $t$ statistic indicates that such a split would lead to two resulting exposure conditions with significantly different  average potential outcomes.

In addition to the $t$ statistic, we can also use weighted SSE as the split criterion. We define 
\begin{equation*}
\begin{split}
\text{Score}_{s}(\mathcal{R}_{\text{left}}, \mathcal{R}_{\text{right}}) = & \text{WSSE}(\mathcal{R}_{\text{left}} \cup \mathcal{R}_{\text{right}}) - \\  & \left(   
\frac{ \left( \sum_i  \mathbbm{1}[\bm{R}_i \in  \mathcal{R}_{\text{left}} ] \right)
\text{WSSE}(\mathcal{R}_{\text{left}}) + \left( \sum_i  \mathbbm{1}[\bm{R}_i \in  \mathcal{R}_{\text{right}} ] \right) \text{WSSE}(\mathcal{R}_{\text{right}})}{\sum_i {\mathbbm{1}[\bm{R}_i \in \left( \mathcal{R}_{\text{left}} \cup \mathcal{R}_{\text{right}} \right) ] } }
 \right), \\
& \text{ where }     \text{WSSE}(\mathcal{R}) = \sum_{i \in \mathcal{N}, \vR_i \in \mathcal{R} } \frac{1}{ \hat{\mathbb{P}}[\bm{R}_i \in \mathcal{R}]  }  (Y_i - \hat{\mu}(\mathcal{R}))^2.
\end{split}
\end{equation*}
\noindent which means the reduction in WSSE after we split $\mathcal{R}$ further into the two disjoint subsets $\mathcal{R}_{\text{left}}$ and $\mathcal{R}_{\text{right}}$ and calculate their average weighted by the number of units in each subset. 
    
    Note that H\'ajek estimator can be derived by minimizing weighted sum of squared errors given a candidate exposure condition, which corresponds to a subset of $[0, 1]^{M}$ (denoted by $\mathcal{R}$). In other words, $\hat{\mu}_{\text{H\'ajek}}(\mathcal{R})$ can be determined by the following equation:
    \begin{equation*}
    \hat{\mu}_{\text{H\'ajek}}(\mathcal{R}) =  \text{arg}\min_{y} \sum_{i \in \mathcal{N}, \vR_i \in \mathcal{R}} \frac{1}{ \mathbb{P}[\bm{R}_i \in \mathcal{R} ]  } (Y_i - y)^2.
    \end{equation*}

\item \textit{(Honest splitting)}  When estimating variance of the tree results,
the algorithm tends to choose a cutoff  such that the objective is minimized; however, it may overfit the training set, leading to an improper cutoff that produces overestimation of the difference between the average potential outcomes of the two child nodes. We thus split all units into training and estimation sets --- the training set is used for tree partitioning only whereas a separate estimation set is used for estimating the mean and variance. This is a common approach to correcting estimations in causal inference \citep{athey2016recursive,kunzel2019metalearners}. 

\end{enumerate}

Algorithm~\ref{algo:tree} details the procedure of the tree-based algorithm.
\begin{algorithm}[h]
\caption{Implementation for the tree-based algorithm}
\label{algo:tree}
\begin{algorithmic}[1]
\linespread{0.9}\selectfont
\Procedure{Split}{$\mathcal{R}$}
    \State $\mathcal{R}^{*}_l, \mathcal{R}^{*}_r = \varnothing$
    \State Compute $\hat{\mu}(\mathcal{R})$ and its variance using estimation set
    \State Initialize $\text{Score}^*$
\For {$m = 1$ \textbf{to} $M$}
        \For {$i \in \mathcal{N}$} \label{random}
            \State $\theta \gets R_{im}$
            \State $\mathcal{R}_{\text{left}} \gets \{j| R_{jm} \leq \theta \And \bm{R}_j \in \mathcal{R} \}$ 
            \State $\mathcal{R}_{\text{right}} \gets \{j| {R}_{jm} > \theta  \And \bm{R}_j \in \mathcal{R} \}$ 
            \If{Eq.~(\ref{eq:pos}) and the splitting rule are satisfied for both $\mathcal{R}_{\text{left}}$ and $\mathcal{R}_{\text{right}}$ }
                \State Compute Score($\mathcal{R}_{\text{left}}$, $\mathcal{R}_\text{right}$) using training set
                \If {Score($\mathcal{R}_{\text{left}}$, $\mathcal{R}_{\text{right}}$) > Score$^*$}
                    \State $\mathcal{R}_{\text{left}}^* \gets \mathcal{R}_{\text{left}}$
                    \State $\mathcal{R}_{\text{right}^*} \gets \mathcal{R}_{\text{right}}$
                    \State $\text{Score}^* \gets  \text{Score}(\mathcal{R}_{\text{left}}, \mathcal{R}_{\text{right}}) $
                \EndIf
            \EndIf
        \EndFor
    \EndFor
    \If {$\mathcal{R}_{\text{left}}^* \neq \varnothing$\text{ and }$\mathcal{R}_{\text{right}}^* \neq \varnothing$} 
        \State \textsc{Split}($\mathcal{R}_{\text{left}}^*$)
        \State \textsc{Split}($\mathcal{R}_{\text{right}}^*$)
    \Else
        \State Define a new exposure condition $\delta$ that corresponds to the subspace $\mathcal{R}$, i.e. $\psi(\delta)=\mathcal{R}$
        \State Add this new exposure condition $\delta$ to the set of all exposure conditions $\Delta$.
    \EndIf
     \EndProcedure
\end{algorithmic}
\end{algorithm}

\subsection{Results for tree-based approach}\label{sec:tree:res}
We first present the results of the tree-based algorithm for this synthetic network in  Figure~\ref{fig:tree_ws}. 
The tree is derived from the estimation set for the honest splitting purpose~\cite{athey2016recursive}.\footnote{We use $t$ statistic and H\'ajek estimator as an illustrative example.} The threshold for $t$ is set to be $1.96$ also for the illustrative purpose only. 
As shown in the upper panel, there are 14 exposure conditions for the Bernoulli randomization. 
We use leaf (i.e. an exposure condition) $\mathcal{R}_5$ as an illustrative example.
If a unit is in control (left split), the fraction of fully-non-treated closed triad is smaller than 51\% (left split), the fraction of fully-treated closed triad is greater than 32\% (right split), and the fraction of treated neighbors with covariate ($X$) equal to 1 is greater than 56\% (right split), 
it specifies a unique exposure condition $\mathcal{R}_5$, which exhibits the largest average potential outcome (3.71) among all exposure conditions under ego being controlled.  
This indicates that even if this unit is in the control group, if the majority of the closely embedded neighborhood is treated, and in particular those neighbors with  $X_j=1$ are also mostly treated, their outcome would be the largest even though they do not receive the treatment. 
This has implications, for instance, for how to utilize this heterogeneous peer effect and network structure to promote more adoptions when the number of treatments is limited. 
The result for cluster randomization yields fewer (five) exposure conditions, as shown in the lower panel of Figure~\ref{fig:tree_ws}.
This is due to the inter-dependence in their treatment assignments as well as the smaller variation in causal network motifs under this randomization. 
Thus, cluster randomization may not always be ideal for revealing interference heterogeneity  as Bernoulli randomization does. 

\begin{figure}
\centering
\includegraphics[width=0.95\linewidth]{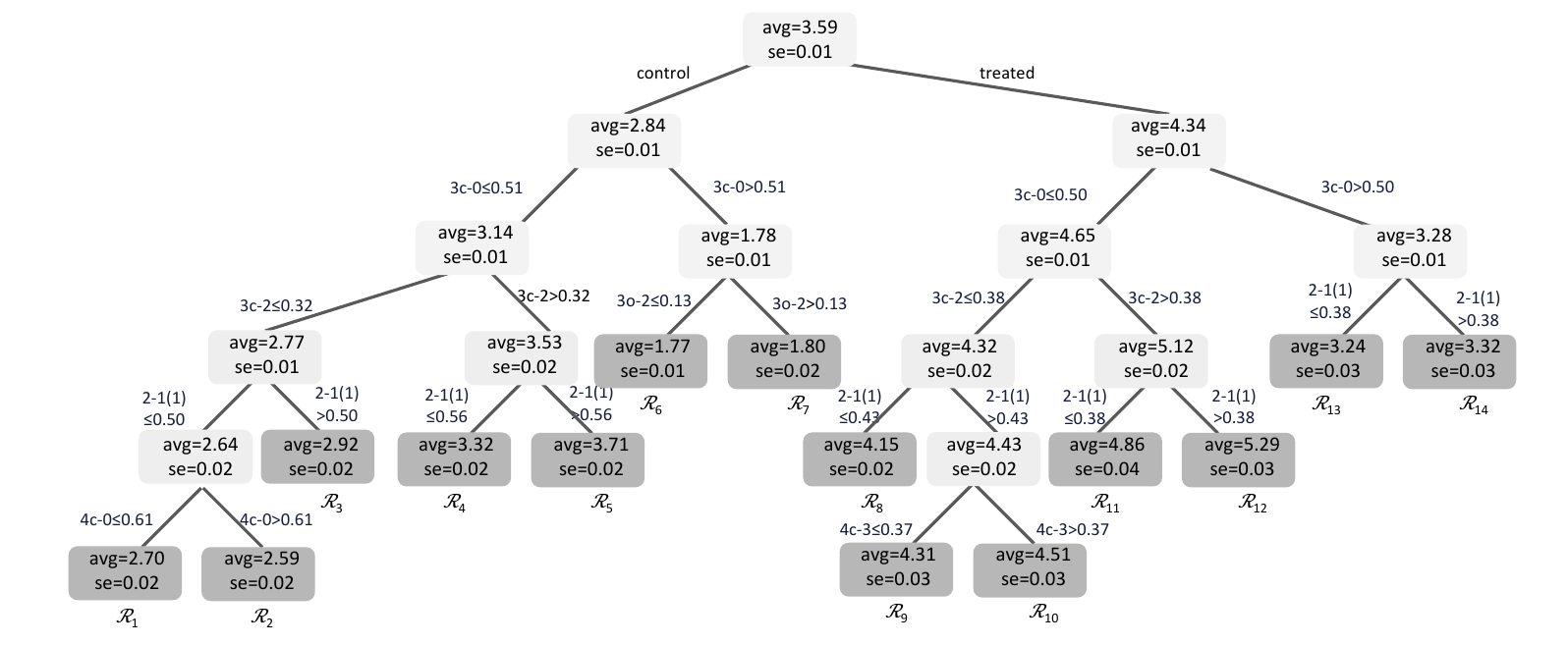}
\includegraphics[width=0.47\linewidth]{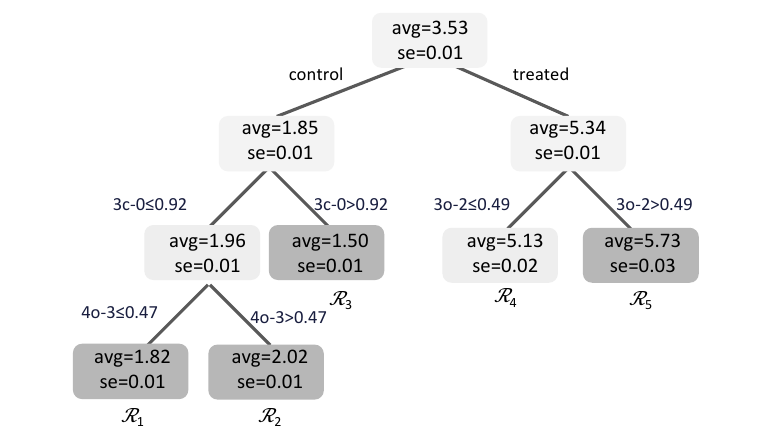}
\caption{\raggedright Tree-algorithm results for the Watts-Strogatz synthetic experiment: (upper) Bernoulli randomization and (lower) graph cluster randomization.
Each leaf corresponds to an exposure condition. The decision path towards each leaf indicates the subspace of $[0,1]^M$ that corresponds to exposure condition. We use training sets to construct the trees and estimation sets to estimate the average potential outcomes and their standard errors. Codes of causal network motifs are marked in Fig.~\ref{fig:causal_network_motif}. ``2-1(1)'' indicates the fraction of treated neighbors with  covariate equal to 1.}
    \label{fig:tree_ws}
\end{figure}

The tree results can also be used to estimate the global average treatment effect -- in this case we need to find the two special exposure conditions that include the fully treated or fully non-treated counterfactual worlds, respectively (in other words, including $\bm{r}^{(0)}$ and $\bm{r}^{(1)}$, respectively).
In the Bernoulli randomization, $\mathcal{R}_{6}$ and $\mathcal{R}_{12}$ contain $\bm{r}^{(0)}$ and $\bm{r}^{(1)}$ respectively, with an average potential outcomes of $5.290 (\pm 0.030)$ and $1.771 (\pm 0.014)$. Note that their difference ($3.519$) is much larger than simply comparing the treatment and the control groups ($4.335-2.842 = 1.493$), and this is a less biased estimation of the global average treatment effect ($4.2$). 
Similarly, for the cluster randomization, 
$\mathcal{R}_3$ and $\mathcal{R}_5$ contain $\bm{r}^{(0)}$ and $\bm{r}^{(1)}$ respectively, with average potential outcomes of $5.732 (\pm 0.026)$ and $1.824 (\pm 0.013)$. This further reduces the bias of estimating the global average treatment effect.

We also compare our approach with the fractional $q$ neighborhood exposure mapping. Using this approach produces four exposure conditions only. 
As shown in Figure~\ref{fig:appendix:WS_qfrac}, less information about heterogeneity of network interference is revealed when only the dyad-level
features are used. For example, 
under Bernoulli randomization, exposure conditions $\mathcal{R}_1$ and $\mathcal{R}_2$ are the only two exposure conditions under control. The fraction of treated neighbors can only detect much smaller differences in the average potential outcomes compared to our approach. 
It also has inferior performance in terms of estimating global average treatment effects. For instance, under Bernoulli randomization, this approach can estimate the global average treatment effect as $4.614-2.380=2.235$, which is much smaller and biased than using the causal network motifs we proposed to use (3.519 for our estimation in Bernoulli case, and 4.200 for the true effect).

\begin{figure}
    \centering
    \includegraphics[width=0.6\linewidth]{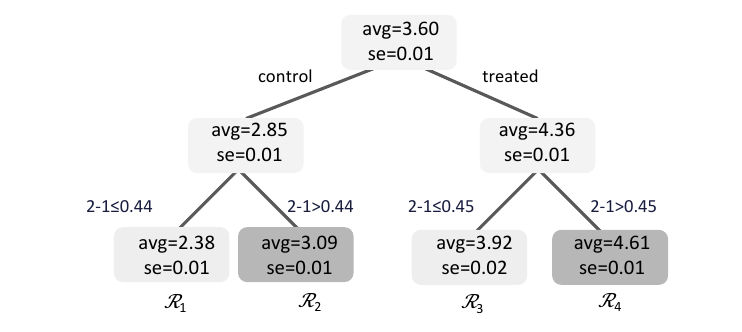}
    \caption{\raggedright Tree-algorithm results for the fractional $q$ neighborhood exposure mapping for Watts-Strogatz synthetic experiment (equivalent to using dyadic features only).
Each leaf corresponds to an exposure condition. The decision path towards each leaf indicates the subspace of $[0,1]^M$ that corresponds to exposure condition. We use training sets to construct the trees and  estimation sets to estimate the average potential outcomes and their standard errors. Codes of causal network motifs are marked in Fig.~\ref{fig:causal_network_motif}. }
    \label{fig:appendix:WS_qfrac}
\end{figure}

\begin{figure}
    \centering\vspace{-1.3cm}
        \includegraphics[width=0.6\linewidth]{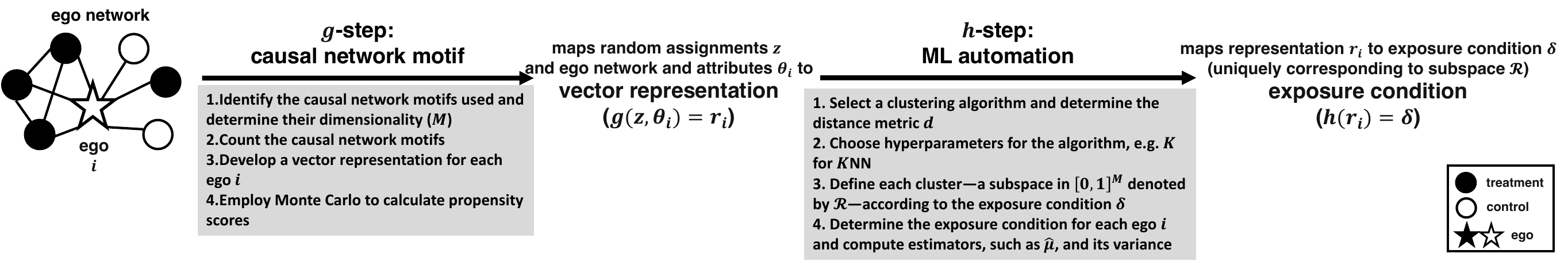}\vspace{-1.2cm}
    \caption{\raggedright Tree-algorithm results for the Instagram test.
Each leaf corresponds to an exposure condition. The decision path towards each leaf indicates the subspace of $[0,1]^M$ that corresponds to exposure condition. We use training sets to construct the trees and estimation sets to estimate the average potential outcomes and their standard errors. Codes of causal network motifs are marked in Fig.~\ref{fig:causal_network_motif}. }
    \label{fig:IG_tree}
\end{figure}

 We also illustrate the tree-based result for the Instagram test in  Figure~\ref{fig:IG_tree}.
From this result, we find a large degree of heterogeneity of network interference  among the control group 
as the control group has four exposure conditions. 
When a user is in the control and their neighbors are mostly in the control group (2-1 $< 27\%$) and a sufficiently large fraction of closed triads are fully non-treated (3c-0 $> 29\%$),  the average potential outcome ($1.11\% \pm 0.04\%$) is the smallest (exposure condition $\mathcal{R}_2$.
By contrast, when the ego is in the control group but $>27\%$ their neighbors are treated and a sufficiently small fraction of closed triads are fully non-treated (3c-0 $\leq 30\%$), the average potential outcome is much larger ($1.24\% \pm 0.04\%$; exposure condition $\mathcal{R}_3$). These results show that even when a person is in the control group, their treated neighbors, especially highly clustered neighbors, still play an important role in prompting them to use it. 
By contrast, for the treatment group, there seems not to be a strong interference pattern.\footnote{The similar average potential outcomes between some exposure conditions, e.g. $\mathcal{R}_5$ and $\mathcal{R}_6$, are due to honest splitting: the estimation from the training set guides such partitioning while being later corrected by the estimation set. }

\section{{Additional Results for WS experiments}}
\label{sec:more}

\subsection{Experimental results for adding more covariates}

In addition to the experimental results in the main text, 
we also perform experiments on the following potential outcomes:
\begin{equation*}
y_i(\vZ) = 1 + Z_i + \sum_{j \in \mathcal{N}_i} 2 w_{ij} Z_j + \frac{1}{5} W_i (1+Z_i) +  \varepsilon_i.    
\end{equation*}
Here, \(W_i = \sum_j \text{cf}_{ij}\), suggesting that if unit \(i\) has many neighbors who are clustered, there is a stronger confounding effect as well as an interaction effect (\(W_iZ_i\)).

We assume that practitioners are unaware that \(W_i=\sum_j \text{cf}_{ij}\) is a key confounder, yet they have access to all network motifs computed during the experiment that are correlated with \(W_i\).

To examine the effect of account for covariates available on estimation, we compare two approaches in Figure~\ref{fig:nei}:
\begin{itemize}
\item \textit{With partialling-out}: We first conduct a regression of \(Y_i\) on all available covariates to remove their effects, which include all counts of network motifs as well as demographic factors. Subsequently, we employ the nearest neighbors method using the residuals of this regression as outcomes and consider only the causal network motifs to compute distances.
    \item \textit{Without partialling-out}: Our standard approach.
\end{itemize}

Using the regression distance metric and the `square-full' causal network motif, we present a comparison of these two approaches. As Figure~\ref{fig:nei} demonstrates, although the estimations with or without controlling for covariates (network motifs) are qualitatively similar, the partialling-out approach substantially reduces estimation variance. Therefore, we recommend that practitioners apply partialling out before utilizing our approach to reduce variance.

\begin{figure}
    \centering
    \includegraphics[width=0.35\linewidth]{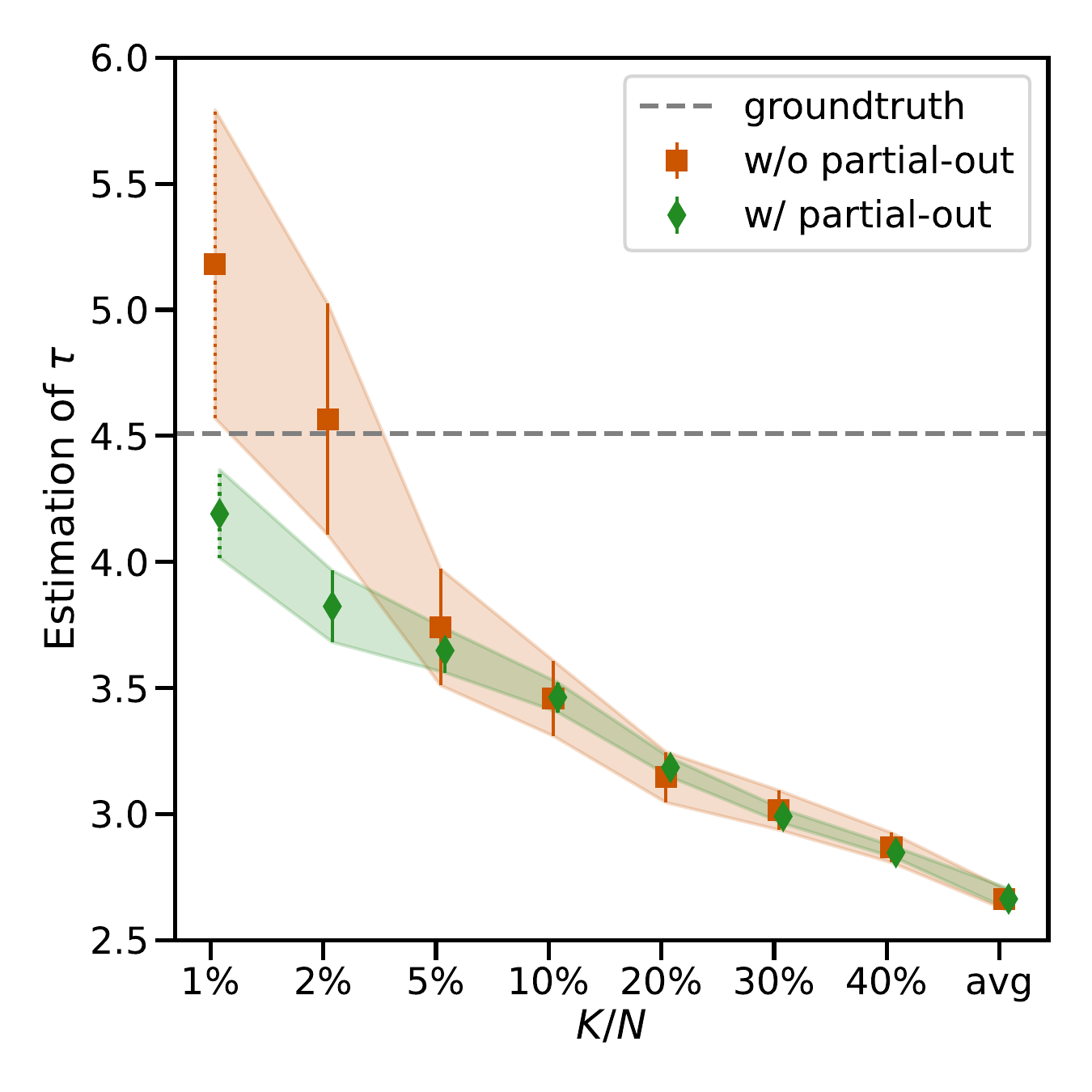}
\caption{\raggedright \scriptsize Nearest Neighbor Estimation Results for the Watts-Strogatz Network with and without partialling out (H\'ajek estimators). The x-label, $K/N$, denotes the fraction of nodes used in the nearest neighbor exposure condition relative to the total population. Error bars represent standard errors, with dotted bars indicating estimates that do not satisfy the positivity requirement. The dashed gray lines depict the ground truth -- the global treatment effect ($\tau$). The label ``avg" refers to the method of calculating average outcomes for treatment and/or control groups. }
    \label{fig:nei}
\end{figure}

\section{Details for the Slashdot Experiment} 
\label{sec:appendix:Slashdot}

Finally, we analyze public network data from Slashdot, a news website renowned for its specific user community, as provided by~\cite{leskovec2009community}. The degree distribution of this network is presented in Figure~\ref{fig:Slashdot:degree}. The network comprises 82,168 nodes and 582,533 edges.

\begin{figure}\centering
\includegraphics[width=0.3\linewidth]{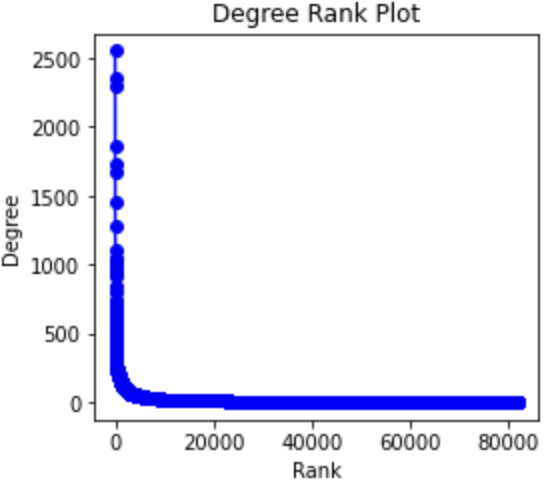}
\caption{Degree rank plot for the Slashdot network}
\label{fig:Slashdot:degree}
\end{figure}

Note that unlike the Watts-Strogatz network, the Slashdot network exhibits a long-tailed degree distribution -- most nodes have small degree whereas a few nodes have a large degree. This long-tailed degree distribution is very common in real-world networks.  We employed the same potential outcome functions as used in the Watts-Strogatz network and applied our methods to this real-world network. For our experiments, both Bernoulli and graph cluster randomizations (with 1024 clusters) were conducted. 
{Here $\mu(\bm{1})=6.0$, $\mu(\bm{0})=1.5$, and $\tau=4.5$}.

Nearest neighbors results are presented in Figure~\ref{fig:appendix:soc}. As illustrated in the figures, all the primary conclusions drawn from the Watts-Strogatz network are applicable here. For example, a smaller value of $K$ results in less biased estimation, albeit with a larger variance. Additionally, integrating our approach with graph cluster randomization yields the least biased estimation. It is worth noting that in comparison to the Watts-Strogatz model, where the degree distribution is less skewed, a smaller value of $K$ seems to violate the positivity requirement more easily in the Slashdot experiment. This is because nodes with a larger degree may be less likely to have a causal network motif representation that closely resembles the fully treated or non-treated representations (much less likely to become a nearest neighbor to $\vr^{(1)}$ or $\vr^{(0)}$).

\begin{figure}
\centering
\includegraphics[width=0.75\linewidth]{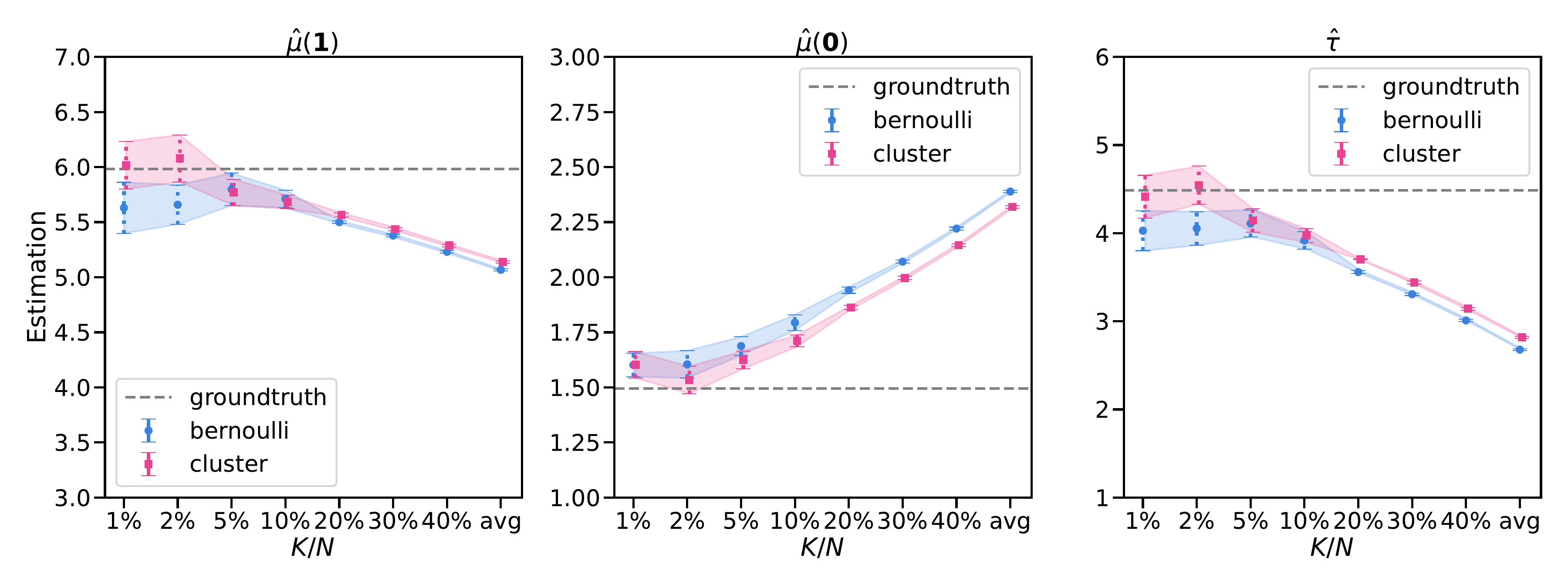}
\caption{\raggedright\scriptsize Nearest Neighbor Estimation Results for the Slashdot Network (H\'ajek estimators). The x-label, $K/N$, denotes the fraction of nodes used in the nearest neighbor exposure condition relative to the total population. Error bars represent standard errors, with dotted bars indicating estimates that do not satisfy the positivity requirement. The dashed gray lines depict the ground truth, signifying the average potential outcomes for fully treated ($\mu(\bm{1})$) or non-treated ($\mu(\bm{0})$) groups, as well as the global treatment effect ($\tau$). The label ``avg" refers to the method of calculating average outcomes for treatment and/or control groups. }
\label{fig:appendix:soc}
\end{figure}

\end{document}